\documentclass{article}
\pdfoutput=1
\usepackage[final,nonatbib]{neurips_data_2023}
\usepackage[numbers]{natbib}

\usepackage{graphicx} %

\usepackage[utf8]{inputenc} %
\usepackage[T1]{fontenc}    %
\usepackage{hyperref}       %
\usepackage{url}            %
\usepackage{booktabs}       %
\usepackage{amsfonts}       %
\usepackage{nicefrac}       %
\usepackage{microtype}      %
\usepackage{multirow}
\usepackage{booktabs}
\usepackage{longtable}
\usepackage{microtype}
\usepackage{graphicx}
\usepackage{booktabs} %
\usepackage{multirow}
\usepackage{makecell}
\usepackage{capt-of}
\usepackage{bm,bbm}
\usepackage[makeroom]{cancel}
\usepackage[table,xcdraw,dvipsnames]{xcolor}
\usepackage{colortbl}
\usepackage{subcaption}
\usepackage{tabularx}
\usepackage{adjustbox}
\usepackage{changes}
\usepackage{tabularx}
\usepackage{hyperref}
\usepackage{pythonhighlight}
\usepackage{color,soul}
\usepackage{xcolor}
\usepackage{xspace}
\usepackage{float} 

\newlength\savewidth\newcommand\shline{\noalign{\global\savewidth\arrayrulewidth
  \global\arrayrulewidth 1pt}\hline\noalign{\global\arrayrulewidth\savewidth}}
\newcommand{\tablestyle}[2]{\setlength{\tabcolsep}{#1}\renewcommand{\arraystretch}{#2}\centering\footnotesize}

\usepackage{pifont}

\newcommand{\nimgdatasets}{\textsc{7}}
\newcommand{\nimginstances}{\textsc{697,435}}
\newcommand{\nimgpatients}{\textsc{258,639}}

\newcommand{\ourmodel}{LLaVA-Rad\xspace}
\newcommand{\ourvit}{BiomedCLIP-CXR\xspace} %
\newcommand{\ourscore}{CheXprompt\xspace}
\newcommand{\eat}[1]{\ignorespaces}

\title{Towards a clinically accessible radiology multimodal model: open-access and lightweight, with automatic evaluation}

\author{Juan Manuel Zambrano Chaves$^{3\diamond}$\thanks{~Equal Contribution. $^\diamond$ Work performed as an intern at Microsoft Research. $^\ddag$ Corresponding authors: swang@cs.washington.edu, hoifung@microsoft.com.}~\, Shih-Cheng Huang$^{3\diamond*}$,  \\
\textbf{Yanbo Xu$^{1*}$, Hanwen Xu$^{2*}$, Naoto Usuyama$^{1*}$, Sheng Zhang$^{1*}$,}\vspace{-2mm}\\\\
\textbf{Fei Wang$^{4}$, Yujia Xie$^{1}$, Mahmoud Khademi$^{1}$, Ziyi Yang$^{1}$, Hany Awadalla$^{1}$,}\\
\textbf{Julia Gong$^{1}$, Houdong Hu$^{1}$, Jianwei Yang$^{1}$, Chunyuan Li$^{1}$, Jianfeng Gao$^{1}$,}\\
\textbf{Yu Gu$^{1}$, Cliff Wong$^{1}$, Mu Wei$^{1}$, Tristan Naumann$^{1}$, Muhao Chen$^{5}$,}\\
\textbf{Matthew P. Lungren$^{1,3,6}$, Akshay Chaudhari$^{3}$, Serena Yeung-Levy$^{3}$, Curtis P. Langlotz$^{3}$,}\\
\textbf{Sheng Wang$^{2,\ddag}$, Hoifung Poon$^{1,\ddag}$}\\\\
$^{1}$Microsoft Research\qquad
$^{2}$University of Washington\qquad
$^{3}$Stanford University\vspace{-1mm}\\\\
$^{4}$University of Southern California\qquad
$^{5}$University of California, Davis\vspace{-1mm}\\\\
$^{6}$University of California, San Fransisco\vspace{-1mm}\\\\
}

\begin{document}

\maketitle
\vspace{-5mm}
\begin{abstract}

The scaling laws and extraordinary performance of large foundation models motivate the development and utilization of such models in biomedicine. 
However, despite early promising results on some biomedical benchmarks, there are still major challenges that need to be addressed before these models can be used in real-world clinics. Frontier general-domain models such as GPT-4V still have significant performance gaps in multimodal biomedical applications.
More importantly, less-acknowledged pragmatic issues, including accessibility, model cost, and tedious manual evaluation make it hard for clinicians to use state-of-the-art large models directly on private patient data. Here, we explore training open-source small multimodal models (SMMs) to bridge competency gaps for unmet clinical needs in radiology. To maximize data efficiency, we adopt a modular approach by incorporating state-of-the-art pre-trained models for image and text modalities, and focusing on training a lightweight adapter to ground each modality to the text embedding space. For training, we assemble a large dataset of over 697 thousand radiology image-text pairs. For evaluation, we propose \ourscore, a GPT-4-based metric for assessing factual accuracy, and demonstrate its parity with expert evaluation. For best practice, we conduct a systematic ablation study on various choices in data engineering and multimodal training. The resulting \ourmodel (7B) model attains state-of-the-art results on standard radiology tasks such as report generation and cross-modal retrieval, even outperforming much larger models such as GPT-4V and Med-PaLM M (84B). Moreover, \ourmodel requires only one day to be trained on over 697 thousand image-text pairs using a standard 8-A100 GPU cluster, allowing further fine-tuning by clinicians using their own data. The inference of \ourmodel is fast and can be performed on a single V100 GPU in private settings, offering a promising state-of-the-art tool for real-world clinical applications.

\end{abstract}

\section*{Introduction}
\label{sec:introduction}

Foundation models, trained on massive amounts of unlabelled data using self-supervised learning, enable rapid adaptation to various downstream tasks with minimal requirement for task-specific labeled data~\cite{bommasani2021opportunities, moor2023foundation,xu2024whole}. Due to the high cost of annotating biomedical data~\cite{huang2023self, huang2022developing}, foundation models are poised to become a new paradigm in biomedicine, achieving state-of-the-art results on many applications, including medical question answering~\cite{medpalm, moor2023foundation} and medical image classification~\cite{shekoofeh2021big, azizi2022robust}. Recently, multimodal generative AI has emerged as an exciting frontier in the biomedical domain, expanding the application scope from single-modality to multi-modality (e.g., text and image), such as visual question answering and radiology report generation~\cite{llavamed, medpalm, tanno2023consensus}. While existing models are still largely evaluated on artificial biomedical benchmarks, their promising performance demonstrates their potential in clinical applications.

However, there are still major bottlenecks hindering the use of foundation models in real-world clinical settings. First, sharing patient data with large foundation models hosted on the cloud is subject to privacy concerns \cite{thirunavukarasu2023large}. Therefore, clinicians may prefer to run inference and fine-tuning locally. Second, existing state-of-the-art models are often very large and resource-intensive, which makes local deployment challenging. Smaller models incur smaller carbon footprint \cite{truhn2023ecological} and offer reduced serving costs and latency, which is of particular importance in resource-constrained settings outside of data centers \cite{wornow2023shaky}. However, while small language models have shown success in text domains \cite{phi, orca2, clint5}, small multimodal models (SMMs) still have significant performance gaps compared to larger models \cite{medpalm, tanno2023consensus}. Third, many state-of-the-art models are inaccessible \cite{wornow2023shaky}, necessitating the development of effective open-source models for biomedicine. Finally, existing evaluation methods of factual correctness exhibit limited correlation with expert assessments \cite{yu2023evaluating}. Since even the best models are still subject to errors such as hallucination, it is crucial to develop reliable methods to evaluate the correctness of model outputs at scale, especially in the specialized field of biomedicine \cite{clusmann2023future}. 

\begin{figure*}[ht]
\centering
\begin{center}
\vspace{-0.02in}
\includegraphics[width=1.0\textwidth]{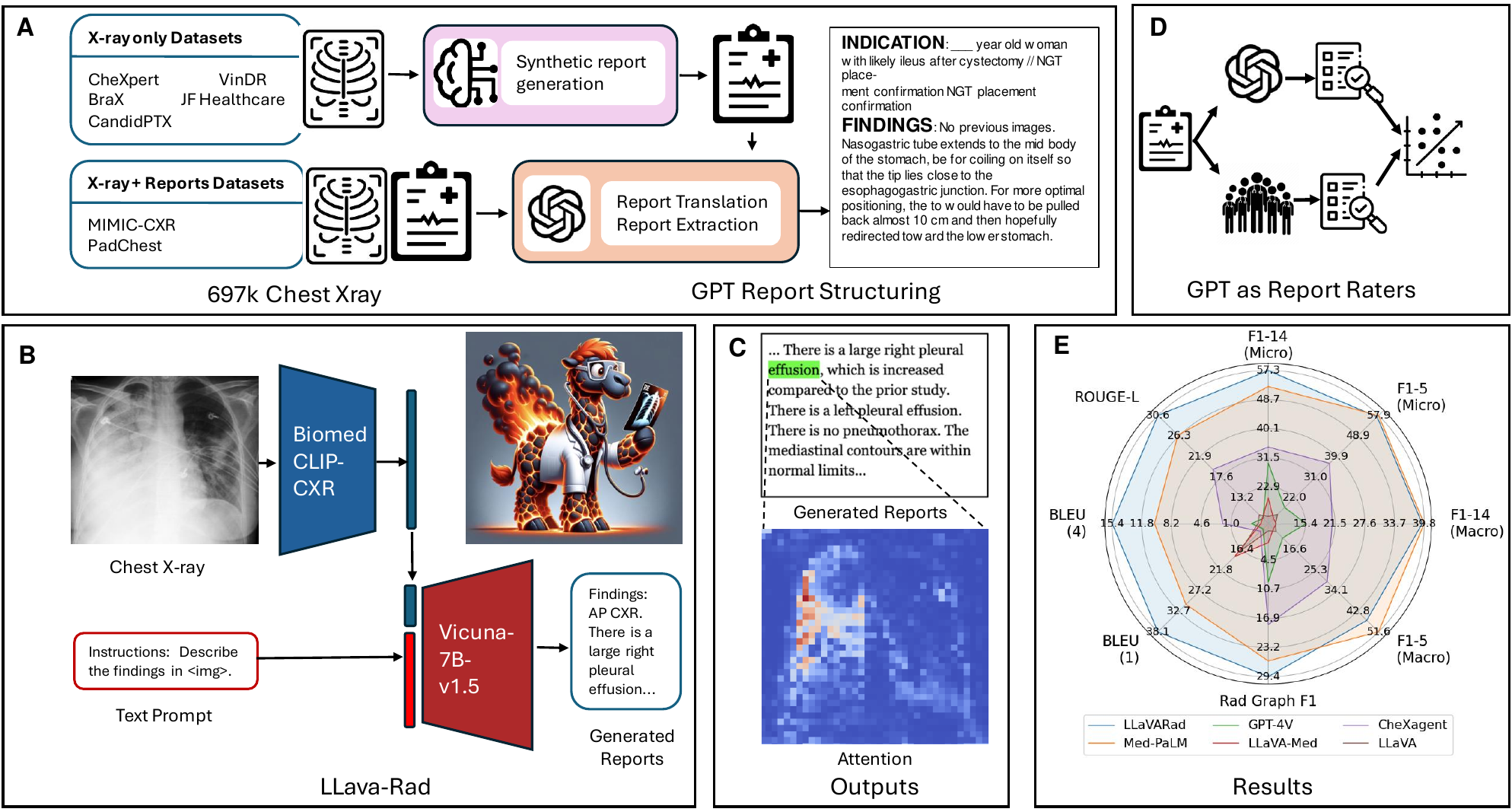}
\end{center}\vspace{-0.01in}
\caption{\textbf{\ourmodel Overview.} \textbf{A}, To train \ourmodel, we assemble a large dataset with over 697 thousand chest X-ray image-text pairs; GPT-4 is used to process and structure the corresponding radiology reports. \textbf{B}, We adopt a modular approach by incorporating state-of-the-art pre-trained models for individual modalities and focusing on training lightweight adapters. \textbf{C}, A qualitative visualization of the model's attention during its generative process. \textbf{D}, For evaluation, we also propose a novel factual error scoring approach using GPT-4 and demonstrate its parity with expert evaluation. \textbf{E}, \ourmodel outperforms much larger models like GPT-4V and Med-PaLM M on prior standard metrics } 
\label{fig:method}\vspace{0.05in}
\end{figure*}

To bridge this gap between existing medical foundation models and real-world clinical applications, we have developed \ourmodel, a SMM that attains state-of-the-art performance in standard radiology imaging tasks (\textbf{Fig. \ref{fig:method}}), in addition to \ourscore, an automatic scoring metric for factual correctness. 
We focus our study on identifying key findings from chest X-ray (CXR) images, the most commonly performed medical imaging examination. Automatically drafting high-quality radiology reports is a challenging but clinically meaningful task that could directly increase radiologist productivity and potentially improve communication and decrease burnout \cite{langlotz2023future}. 
Existing frontier models such as GPT-4V still have a large performance gap even on such a fundamental medical application. 
To develop \ourmodel, we adopt a modular approach by incorporating state-of-the-art open-source pretrained models for image and text modalities, and focusing on training a lightweight adapter to ground each modality to the text embedding space. 

For training, we assemble a large dataset comprising {\nimginstances} radiology image-report pairs from {\nimgdatasets} diverse sources. Some data sources only contain structured labels of key findings, in which case we use GPT-4 to synthesize the report based on the ground-truth labels. We only apply this GPT-4 synthesis to the training data. 

For evaluation, we report standard metrics such as $n$-gram-based BLEU and ROUGE, as well as factuality checks based on CheXpert and RadGraph \cite{smit2020chexbert,jain2021radgraph}. Additionally, we propose \ourscore, a factuality evaluation method based on GPT-4. Compared to existing automatic metrics, we show that \ourscore is more consistent with error quantification by practicing radiologists, thus demonstrating the potential of using GPT-4 for evaluation in a manner that is both scalable and highly relevant to medical practice. 

To establish best practices for biomedical multimodal learning, we conduct a systematic ablation study on various choices in data engineering and multimodal training. The resulting \ourmodel (7B) model attains state-of-the-art results on standard radiology tasks such as report generation and cross-modal retrieval, even outperforming much larger models such as GPT-4V and Med-PaLM M (84B)~\cite{medpalm}.

\ourmodel inference is fast and can be run on a single V100 GPU in private settings, offering a promising state-of-the-art tool for real-world clinical applications. In addition, \ourmodel training is also very efficient, taking just one day on over 697 thousand image-text pairs using a standard 8-A100 cluster. This means that clinicians can further efficiently fine-tune the model as needed using their private data. By examining the model weights, we found that \ourmodel can ground key regions of abnormalities to generated words in the output report, which signifies future opportunities to synergize with latest progress in biomedical segmentation and grounded report generation. 

In summary, we develop \ourmodel, a lightweight yet high-performing radiology multimodal model for clinical applications. The promising performance of \ourmodel shows that its underlying modular approach can effectively and efficiently bridge the multimodal performance gap in existing frontier models, enabling clinical access with limited computational resources. 

\section*{Results}
\label{sec:results}
\subsection*{Overview of \ourmodel}
\ourmodel represents an emerging paradigm in exploring small multimodal models (SMMs), following the proliferation of small language models (SLMs) (\textbf{Fig. \ref{fig:method}}). Despite its compact size, which is over an order of magnitude smaller than prior state-of-the-art models such as Med-PaLM M, \ourmodel attains state-of-the-art performance on standard radiology tasks. This bodes well for potential clinical applications with limited computational resources.

Our intuition for designing \ourmodel is that a lightweight, specialized SMM can be efficiently developed by decomposing training into unimodal pretraining on individual modalities followed by lightweight cross-modal learning focusing on a small adapter to ground a non-text modality to the text embedding space. For training, we collect 697 thousand pairs of fully de-identified CXR images and associated radiology reports from {\nimgdatasets} diverse datasets \cite{irvin2019chexpert, reis2022brax, Feng2021, nguyen2020vinbigdata, jfhealthcare, johnson2019mimic, bustos2020padchest}. These de-identified image-text pairs were sourced from approximately {\nimgpatients} patients. 
The diversity of this data facilitates the pretraining of robust unimodal models (image encoder in this case) and cross-modal adapters (grounding image to text).

\ourmodel can generate radiology report findings given a CXR image. Its training comprises three stages: a pre-training stage, an alignment stage, and a fine-tuning stage. In the first stage of pre-training, we train a domain-specific vision encoder by using 697 thousand pairs of CXR images and associated radiology reports. Since CXR images are often published with a limited number of associated findings or image labels instead of a complete report, we used GPT-4 to synthesize a report based on annotated image labels. Alternatively, reports may be available in other languages, such as the PadChest reports which are available in Spanish, for which we leverage GPT-4 to translate them into English. We also exploit GPT-4 to extract findings from reports when the finding section cannot be reliably extracted using existing rule-based heuristics~\cite{johnson2019mimic}. These diverse datasets offer us a robust and effective vision encoder for embedding CXR images with the consideration of the associated radiology reports. In the second stage of alignment, we align this pre-trained vision encoder with radiology report findings by training a conditional generative model that generates findings given an input CXR. In this alignment stage, we provide only the CXR as the input, without any associated contexts such as clinical instructions or patient information. As noted by other works~\cite{liu2023improvedllava,llavamed} and also demonstrated by our ablation studies, this strategy can substantially improve the alignment by forcing the decoder model to focus on the image alone. In the third stage, we fine-tune the model to generate the findings given both the indication for the exam and the image, more closely reflecting the real-world setting. \ourmodel exploits an efficient technique LoRA \cite{hu2021lora} for fine-tuning, thus substantially reducing the computational time required for this stage. We further reduce the computational time by only using MIMIC-CXR training data instead of the entire 697 thousand image-text pairs in the second and third stages, since reports in MIMIC-CXR are of higher quality. The three stages of \ourmodel can be finished in 8 hours, 4 hours, and 16 hours respectively using 8 A100 GPUs.

\subsection*{Evaluating \ourmodel using existing report generation benchmarks}

\begin{figure}[H]
\begin{center}
\includegraphics[width=1.0\textwidth]{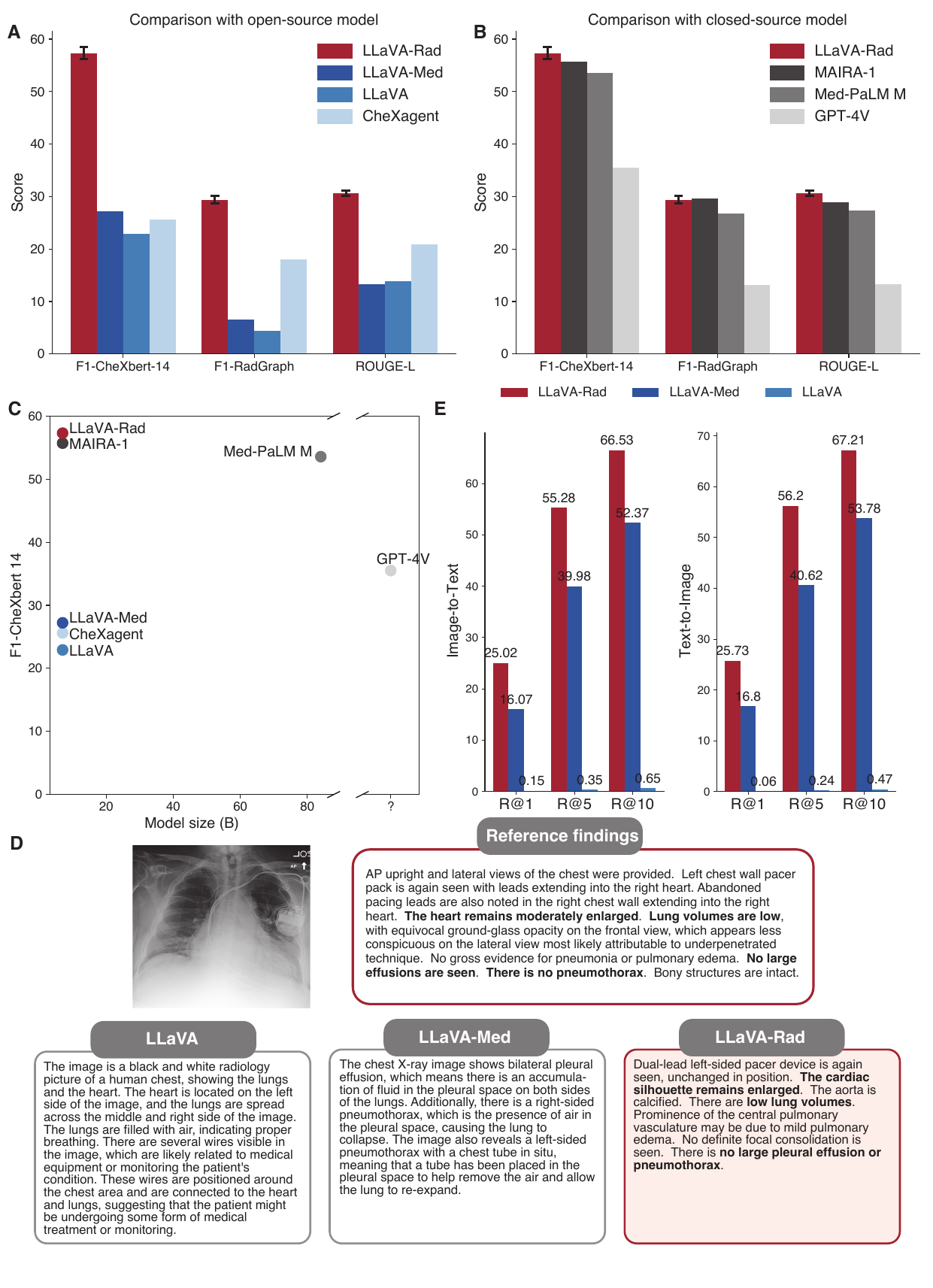}
\end{center}\vspace{-0.1in}
\caption{\textbf{Quantitative and qualitative evaluation of \ourmodel using existing report generation benchmarks.} \textbf{A}, Comparison between \ourmodel and open-source models according to existing preferred factual correctness and lexical similarity metrics. \textbf{B}, Comparison between \ourmodel and closed-source models according to existing preferred factual correctness and lexical similarity metrics. \textbf{C}, Comparison between model size and factual correctness shows that \ourmodel is both smaller and more factually correct compared to existing approaches. \textbf{D}, Illustration of a sample generated report from \ourmodel compared with that of LLaVA and LLaVA-Med. \ourmodel's generations that match reference findings are highlighted. \textbf{E}, Comparison of the performance on cross-modal retrieval demonstrated by \ourmodel, LLaVA-Med and LLaVA. } 
\label{fig:fig2}\vspace{0.05in}
\end{figure}

We first evaluated \ourmodel on the widely-used radiology report generation benchmark MIMIC-CXR (\textbf{Fig. ~\ref{fig:fig2}, Supplementary Figure ~\ref{fig:breakdown}}) using metrics assessing lexical similarity and factual accuracy. In particular, lexical similarity metrics, such as BLEU and ROUGE, are used in traditional NLP to assess the model's ability to produce contextually and stylistically aligned output. On the other hand, factual accuracy metrics, including CheXbert-based and RadGraph-based F1 scores~\cite{smit2020chexbert,jain2021radgraph} are more clinically relevant because they gauge the extent to which the generated reports accurately reflect imaging findings.

We found that \ourmodel achieved superior performance on both groups of metrics  (\textbf{Fig. \ref{fig:fig2}A-D}). When compared to other models of equivalent size (7B parameters), such as LLaVA-Med~\cite{llavamed}, CheXagent~\cite{chen2024chexagent} and MAIRA-1~\cite{hyland2023maira}, \ourmodel demonstrates significant advancements in performance across all evaluated metrics. Furthermore, \ourmodel is more efficient than the current overall leading model, Med-PaLM M~\cite{medpalm}, with an order of magnitude fewer parameters. This efficiency does not come at the cost of effectiveness; \ourmodel outperforms Med-PaLM M in the most important existing lexical similarity and factual correctness metrics for radiology text (ROUGE-L and F1-RadGraph \cite{rales}, with a relative improvement of 12.1\% and 10.1\% respectively). A more detailed evaluation (\textbf{Supplementary Table \ref{tab:main-results}}) shows that Med-PaLM M marginally surpasses \ourmodel by F1-5 CheXbert metrics, which assess only a small subset of 5 potential abnormalities, and the performance gap is minimal (<1\% relative improvement). Most of these competing models also use MIMIC-CXR for training (with the notable exception of LLaVA-Med). We attribute the promising performance of \ourmodel to its modular design, which is more data efficient. The efficiency and the high degree of factual and lexical precision of \ourmodel demonstrate its potential in real-world applications where large models are computationally too costly.

\begin{figure}[H]
\begin{center}
\includegraphics[width=1.0\textwidth]{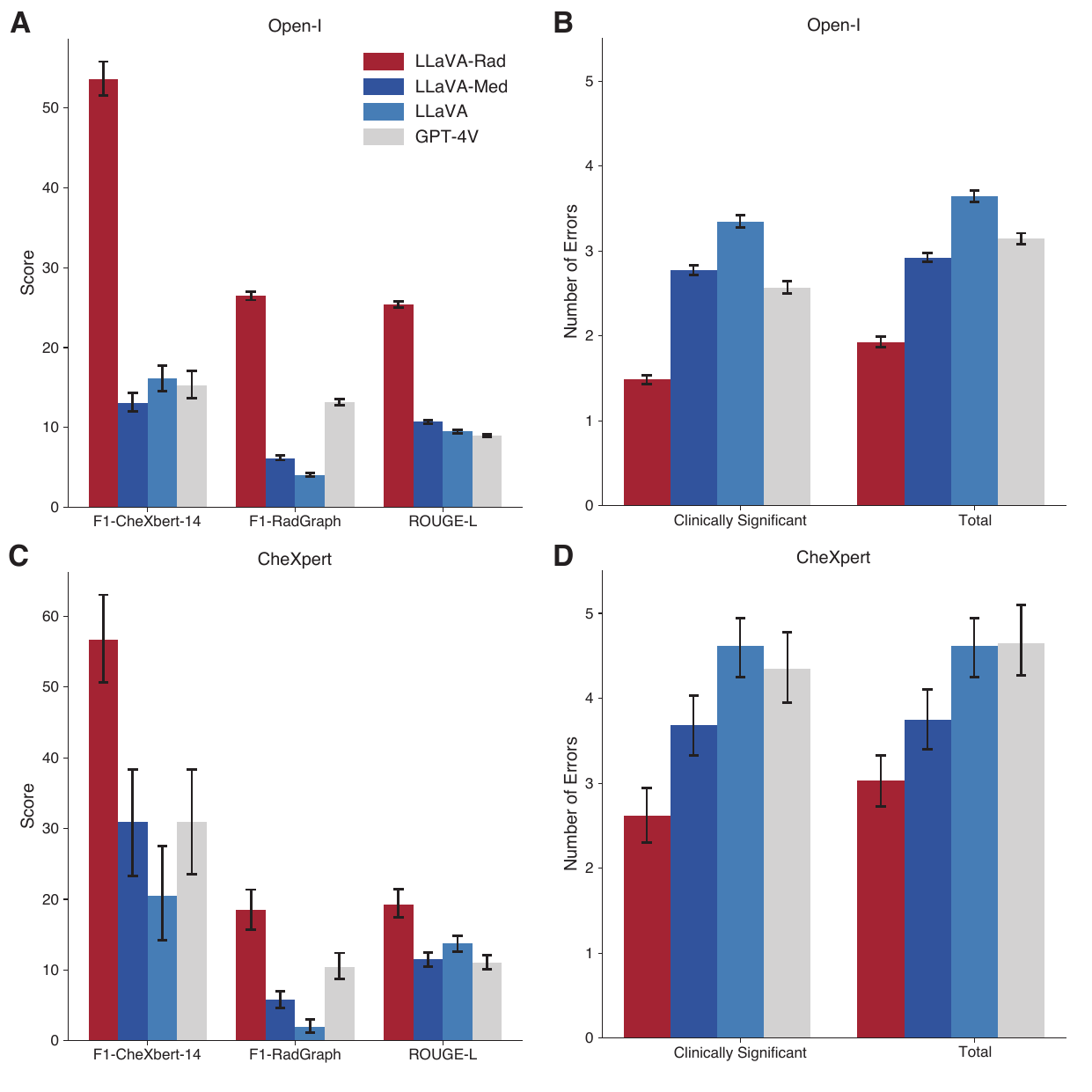}
\end{center}\vspace{-0.2in}
\caption{External validation results for \ourmodel on held out Open-I (n=2,163, panels \textbf{A} and \textbf{B}) and CheXpert (n=63, panels \textbf{C} and \textbf{D}) datasets.} 
\label{fig:fig4}
\end{figure}

The results of our evaluations reveal that \ourmodel's superior performance is consistent across other datasets, including CheXpert~\cite{irvin2019chexpert} and Open-I~\cite{demner2012design}. Notably, CheXpert was used solely for pretraining the image encoder, while Open-I was entirely new to the model, underscoring its robustness and adaptability. Similar to the evaluation on MIMIC, We also employ CheXbert-14, F1-RadGraph, and ROUGE-L to assess the factual accuracy and lexical similarity of the reports on these datasets. As illustrated in \textbf{Fig.~\ref{fig:fig4}}, \ourmodel significantly outperforms LLaVA-Med, LLaVA, and GPT-4V across all metrics on these datasets.

Finally, to verify the effectiveness of our idea in generating aligned vision and language representations, we examined the learned image encoder in \ourmodel by comparing the performance of using it for retrieval against the image encoders from LLaVA and LLaVA-Med. We observed that \ourmodel attained the best results on both image-to-text and text-to-image retrieval, indicating the high quality of its image encoder by training on 697 thousand text-image pairs (\textbf{Fig. ~\ref{fig:fig2}E}). Moreover, LLaVA-Med performed better than LLaVA, suggesting that better performance can be gained as increasing domain specialization is performed.

\subsection*{Evaluating \ourmodel using \ourscore, a GPT-4-based evaluation system}
It is well known that existing ngram-based automated report evaluation methods might be biased to pre-defined conditions and have limited correlation with expert assessments \cite{yu2023evaluating}. We thus explore the utility of an LLM-based evaluation system, which has shown success in other domains \cite{liu2023gpteval, wang2023chatgpt, gilardi2023chatgpt}. Specifically, we employ GPT-4 as an evaluator to count how often the generated report contains errors in each of the following six categories, as per a previous study \cite{yu2023evaluating}: false positive finding, omission of finding, incorrect location/position of finding, incorrect severity of finding, mention of comparison that is not present in the reference report, omission of comparison describing a change from a previous study. For each error type, we further instruct GPT-4 to distinguish clinically significant and clinically insignificant errors.

\begin{figure}[H]
\begin{center}
\includegraphics[width=1.0\textwidth]{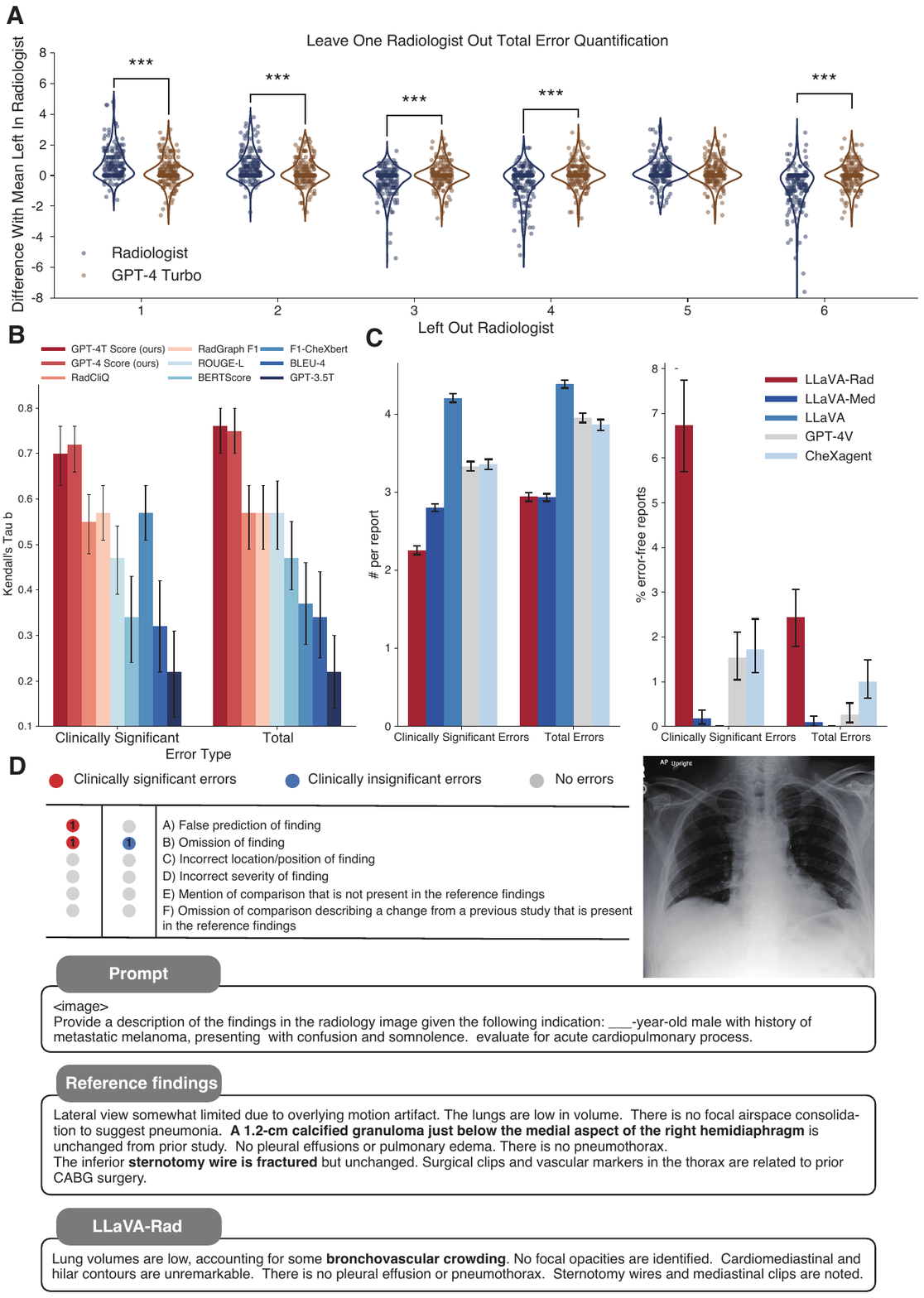}
\end{center}\vspace{-0.01in}
\caption{\textbf{Evaluating \ourmodel using \ourscore.} 
GPT-4T stands for GPT-4 Turbo. 
\textbf{A}, GPT-4 based \ourscore is more similar to average left-in radiologists in total error quantification, compared to the left-out radiologist (mean absolute difference 0.55 vs 0.71). \textbf{B}, Comparison between \ourscore and existing metrics in terms of agreement with radiologist error quantification. \textbf{C}, Comparison between \ourmodel and competing methods using \ourscore on the MIMIC-CXR test set. \textbf{D}, Illustration of how \ourscore can be used to evaluate a report generated by \ourmodel, with errors highlighted.}
\label{fig:fig3}
\end{figure}

We first assessed the rigor of \ourscore by examining its consistency with expert scoring. To this end, we exploited the ReXval dataset \cite{yu2023radiology}, which contains annotations from 6 board certified radiologists on 200 pairs of ground-truth reports from MIMIC-CXR and AI-generated reports. Each radiologist annotates the aforementioned errors in the generated report, also discriminating between clinically significant and insignificant errors. We found that GPT-4-based evaluations were highly correlated with expert scoring by achieving Kendall's Tau-b correlations greater than 0.75 for total errors and greater than 0.70 for clinically significant errors (\textbf{Fig.~\ref{fig:fig3}B}). In contrast, none of the existing preferred metrics (ROUGE-L metric, RadGraph-F1, and RadCliQ) obtained a correlation greater than 0.57. Moreover, we found that a similar evaluation system using GPT-3.5 Turbo, a less capable model compared to GPT-4, attains a much lower association with expert scoring due to an overestimation of the number of total and clinically significant errors, demonstrating the difficulty of automatically scoring radiology reports.

We also perform a head-to-head comparison of the calculation of total errors as determined by GPT-4 Turbo with manual expert ratings in a leave-one-rater-out fashion. \textbf{Fig.~\ref{fig:fig3}A} summarizes the results of this comparison, which on average shows a mean absolute difference (MAD) of 0.71 between the left-out rater and the average of the remaining ones, whereas GPT-4 Turbo has on average 0.55 MAD. We find that the MAD between GPT-4 Turbo \ourscore score and the left-in expert average is smaller compared to the left-out expert in 3 out of 6 cases (P<0.001), and not significantly different (P>0.05) in the remaining 3 out of 6 cases. Altogether, we find that GPT-4 Turbo is indistinguishable from expert raters in calculating the total number of errors, increasing our confidence in using this proposed automated metric as a new evaluation method that directly aligns with expert opinions.

After assuring the effectiveness of the GPT-based metric, we evaluated the performance of \ourmodel on the held-out MIMIC-CXR test set using \ourscore (\textbf{Fig.~\ref{fig:fig3}C,D}). Similar to our observation using existing metrics, \ourmodel outperforms publicly available report generation models, generating fewer clinically significant and total errors compared to GPT-4V and CheXagent. Moreover, by comparing models within the LLaVA family (e.g., \ourmodel, LLaVA-Med, LLaVA), we observed that fewer errors are made in the generated reports as increasing domain specialization is performed. In particular, \ourmodel generates fewer errors than LLaVA-Med, a LLaVA model tailored to medicine, and LLaVA-Med generates fewer errors than the general-domain model LLaVA. This suggests a trade-off between domain-specific performance and broad applicability, supporting our intuition of developing \ourmodel by continual-pretraining of a general model using large amounts of domain-specific data. Finally, to determine the clinical utility of \ourmodel, we explore using the percentage of error-free reports to track the overall performance of report-generation models. A higher percentage of error-free reports increases the utility of a report generation model, given that it directly reflects the number of reports that require little to no radiologist modification following automated generation. Notably, \ourmodel has the highest percentage of error-free reports, with 6.79\% reports free of clinically significant errors, and 2.58\% free of errors. The same trend of improved performance of \ourmodel was observed in the external validation datasets (CheXpert and Open-I), where LLaVA-Rad demonstrated fewer clinically significant and total errors (\textbf{Fig.~\ref{fig:fig4}B,D}), with up to 26\% error-free reports. \ourscore illustrates that while there undoubtedly remains room for improvement in automated radiology report generation, the improvement demonstrated by our model is promising.

\subsection*{Analyzing components of \ourmodel using ablation and case studies}

\begin{figure}[H]
\begin{center}
\includegraphics[width=0.95\textwidth]{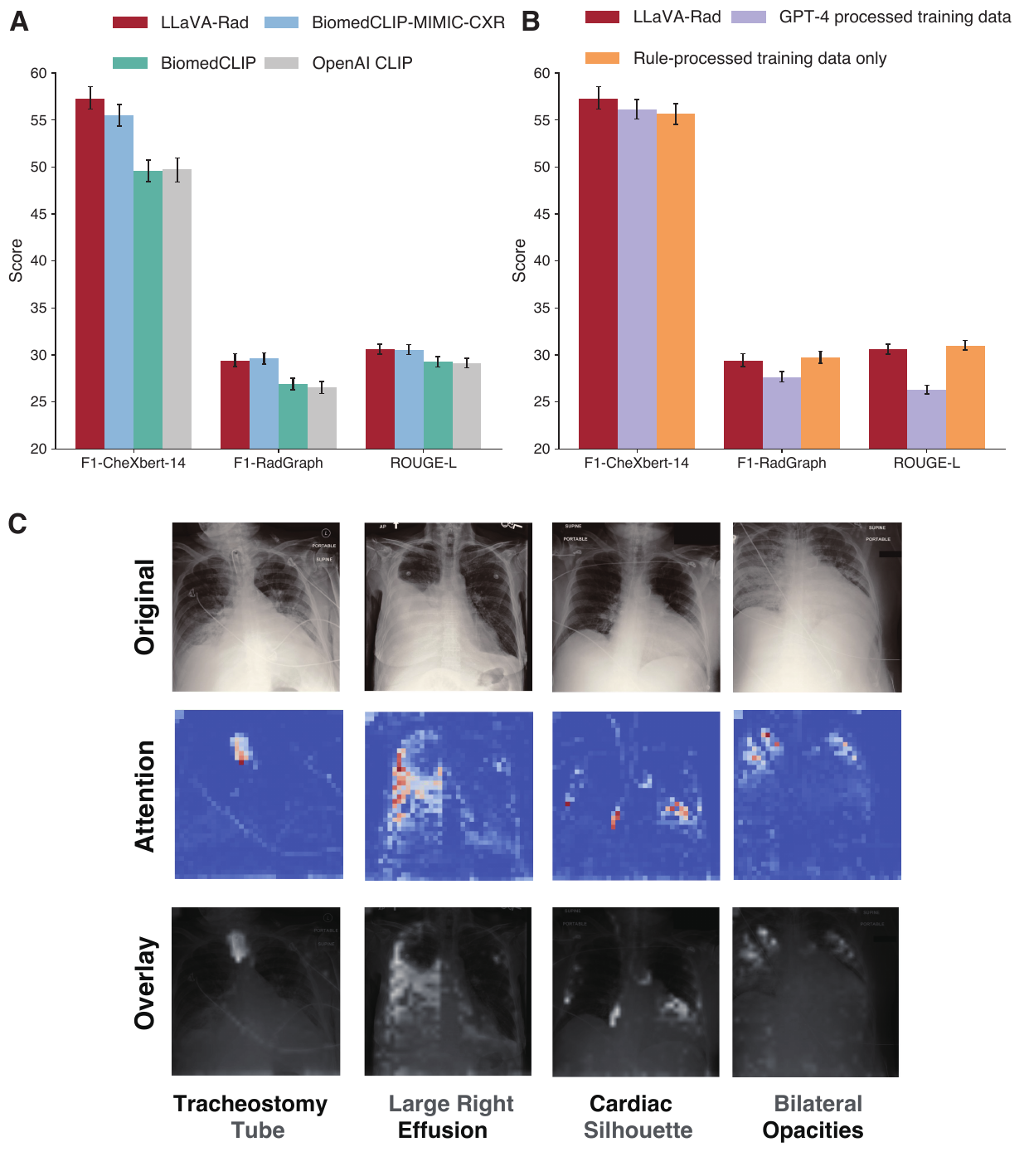}
\end{center}
\caption{\textbf{Analyzing the performance of \ourmodel using ablation studies and attention visualization.} \textbf{A}, Comparison of using different image encoders (from \ourmodel (\ourvit), BiomedCLIP continually pre-trained on MIMIC-CXR, BiomedCLIP, and OpenAI CLIP) to start the alignment and fine-tuning stages. \textbf{B}, Ablation study on only using rule-processed MIMIC-CXR training data or GPT-4 processed training data in alignment and fine-tuning stages. \textbf{C}, Attention visualization qualitatively demonstrates the appropriate grounding of \ourmodel in specific image regions when generating a word (\textbf{bold}) as part of a specific finding (bottom row).} 
\label{fig:abl_atten}
\end{figure}

Conducting thorough ablation studies for LLMs is often intractable due to the costly training of multiple variants. In contrast, the small size of \ourmodel enables us to efficiently conduct ablation studies that explain the promising performance of \ourmodel and potentially inform design choices for larger models. We compared \ourmodel with 8 variants described in \textbf{Supplementary Table \ref{tab:ablation}}. In particular, we investigate two key technical ideas used in \ourmodel: the effect of pre-training a domain-specific image encoder using 697 thousand diverse CXR image-text pairs (\textbf{Fig. ~\ref{fig:abl_atten}A}) and the effect of using GPT-4 to augment and organize the data (\textbf{Fig. ~\ref{fig:abl_atten}B}). First, to understand the effect of pre-training an image encoder, we compare \ourmodel with three increasingly domain-specific variants: an image encoder from OpenAI CLIP, an image encoder using BiomedCLIP, and an image encoder from BiomedCLIP but continually pre-trained using MIMIC-CXR only (\textbf{Fig. ~\ref{fig:abl_atten}A}). To avoid data contamination, we only used the training split of MIMIC-CXR. We did not find noticeable overlap between MIMIC-CXR training split and the PubMed data used to pre-train BiomedCLIP. We found that the MIMIC-CXR-based image encoder outperforms the other two variants, indicating the effectiveness of training a domain-specific image encoder. \ourvit outperforms the BiomedCLIP image encoder continue-pretrained only on MIMIC-CXR (BiomedCLIP-MIMIC-CXR), illustrating the advantage in pre-training using more diverse CXR datasets. Second, we studied the effect of using GPT-4 to process and augment the MIMIC-CXR report data (\textbf{Fig. ~\ref{fig:abl_atten}B}). \textbf{Supplementary Table \ref{tab:mimic_number}} summarizes the data that \ourmodel uses for training in the second and the third stages. It is a combination of rule-based and GPT-structured data. We compare \ourmodel with a variant that only uses rule-based data and a variant that only uses GPT-structured data. We found that \ourmodel attains a better performance than both variants, indicating the effectiveness of GPT-4 data augmentation. The variant that only uses GPT-structured data outperforms the one that only uses rule-based data on factual accuracy metrics, confirming the effectiveness of GPT-4-based structuring in generating clinically precise reports. Finally, it is expected that rule-based variant outperforms GPT-structured variant on n-gram lexical metrics, because the test data is also from rule-based data. These ablation studies support our intuition that domain-specific data can help us build a small but effective domain-specific model, and help inform best practice in training larger models.

We also investigate how \ourmodel's attention map on the input image correlates with a given generated word in the report (\textbf{Fig. ~\ref{fig:abl_atten}C}), which demonstrates the model's ability to focus on relevant image regions for the generation. A detailed examination reveals a significant variability in attention across different layers and attention heads, with different configurations gravitating towards distinct regions of the image (\textbf{Supplementary Figures ~\ref{fig:attn-effusion},~\ref{fig:attn-opac},~\ref{fig:attn-devices},~\ref{fig:attn-aortic},~\ref{fig:attn-opac-2}}). Our evaluation also identifies that the aggregation of attention, particularly through averaging the outputs of all heads within the 20th layer, generally yields the most coherent and relevant focal points across a wide array of scenarios. However, this approach does not uniformly apply, as deviations in alignment were observed in certain instances. Conversely, an alternative strategy of taking the maximum across all layers, coupled with an average across heads, demonstrates a consistently high correlation with pertinent image regions. Our proposed attention visualization indicates a strong alignment between the model's attention and the specific image regions relevant to the generated words. This alignment underscores the model's efficacy in synthesizing contextual information from visual cues to ground its linguistic output.

\section*{Discussion}
\label{sec:discussion}

To address the significant challenges of developing foundation models for real-world clinical settings, our work introduces \ourmodel, a lightweight radiology SMM that offers open-source accessibility while attaining new state-of-the-art results in the domain of radiology report generation. By curating a dataset of 697 thousand CXR images paired with radiology reports from diverse sources, using GPT-4 for dataset processing, coupled with a modular three-stage curriculum training method, we have developed a model that outperforms its larger counterparts, such as GPT-4V and Med-PaLM M, and demonstrates exceptional proficiency in generating accurate and lexically similar radiology reports on the evaluation datasets. Through our attention visualization techniques, \ourmodel offers deep insights into how it prioritizes key regions in chest X-rays, correlating them with specific findings in the generated reports. Furthermore, our work introduces \ourscore, which successfully resolves a major bottleneck of automatic evaluation of factual accuracy of generated radiology reports, by not only demonstrating a closer alignment of automated scoring compared to traditional metrics, but also exhibiting performance on par with expert radiologist annotators. This improved evaluation further illustrates \ourmodel's superiority in clinical report generation.

The landscape of AI-driven radiology report generation has evolved significantly with the advent of transformers and large multimodal models, ushering in a new era of more sophisticated and accurate models~\cite{mohsan2022vision, hyland2023maira, wang2022cross, pan2023s3, hou2023mkcl, huang2023generative, aksoy2023radiology}. R2Gen stands out as a pioneering effort in leveraging memory-efficient transformers for report generation~\cite{chen2020generating}. A notable leap forward is CheXagent~\cite{chen2024chexagent}, which leverages an instruction fine-tuned foundation model trained across 28 publicly available datasets, demonstrating an enhanced capability for analyzing and summarizing CXR images. Concurrently, Flamingo-CXR fine-tuned the Flamingo vision language model~\cite{alayrac2022flamingo} and incorporated regularization and adaptation techniques to tailor their applications to the nuances of radiology report generation~\cite{tanno2023consensus}. 
Med-PaLM M pushed the boundaries by creating a versatile 84-billion-parameter biomedical AI system capable of addressing multiple tasks across various medical modalities~\cite{medpalm}. In contrast to these advancements, our method, \ourmodel, distinguishes itself by not only achieving superior performance across several benchmark metrics but also by being comparatively lightweight. This attribute is particularly important, as it offers a more accessible and efficient solution for scaling radiology report generation, addressing both the need for factual correctness and the practicality of deployment in clinical settings.

While \ourmodel represents a substantial advancement in radiology multimodal models, our research acknowledges several areas for future exploration and improvement. First, the current scope of \ourmodel is limited to CXRs. While CXR is the most common medical image examination, future iterations should evaluate the feasibility of our method on alternative anatomies (e.g., abdomen or extremities) and modalities (e.g., computed tomography or ultrasound) to enhance the model's applicability and utility across diverse application scenarios. 
Attention-based attribution methods have been found to be more effective at explaining model decisions and to be more useful by radiologists \cite{wollek2023attention}, and our attention visualization technique does appear to highlight sensible patterns. However, alternative saliency-based methods for CXR interpretation algorithms such as Grad-CAM have been shown to have limited correlation with expert assessments and limited robustness to input perturbations \cite{saporta2022benchmarking, zhang2024revisiting}. There is a pressing need for a more exhaustive evaluation of such grounding strategies. These would further improve the model's explainability and interpretability, making it more transparent and trustworthy for clinical use. Another consideration is the inherently multimodal nature of modern medical practice, which integrates various patient information streams, including historical medical images, medical records, lab tests, and vital signs. Integrating these diverse and longitudinal data sources into medical multimodal models like \ourmodel could significantly enrich the model's understanding and analysis, leading to more nuanced and holistic patient assessments.

\ourmodel exemplifies a significant leap toward making advanced diagnostic capabilities accessible with limited computational resources, thus paving the way for broader clinical applications and impact. The pursuit of open-source, lightweight, high-performing models that not only extends to various medical imaging types but also incorporates multimodality and interpretability, embodies the next frontier in medical multimodal model development. Such advancements will bridge the gap between current technological capabilities and the real-world demands in clinical applications, moving us closer to achieving meaningful improvements in patient outcomes.

\newpage
\section*{Methods}
\label{sec:methods}

\subsection*{Details of the dataset}
\textbf{CXR-697K} We compiled a comprehensive dataset comprising 697 thousand pairs of CXR images, each accompanied by its corresponding radiology report, for pre-training the image encoder of \ourmodel. This dataset amalgamates data from seven publicly available datasets as summarized in \textbf{Supplementary Table \ref{table:dataset}}. To maintain transparency and reproducibility, we adhere to the original train/val/test splits provided by each contributing public dataset, using only the train split for pre-training the image encoder. 

The CheXpert dataset~\cite{irvin2019chexpert} consists of retrospectively collected chest radiographic studies conducted between October 2002 and July 2017, encompassing both inpatient and outpatient centers at Stanford Hospital. BraX~\cite{reis2022brax}, obtained from chest radiography studies at Hospital Israelita Albert Einstein (HIAE) before the COVID-19 pandemic in S$\tilde{a}$o Paulo, Brazil, was labeled for 14 radiological findings using the CheXpert Label Extraction Algorithm~\cite{irvin2019chexpert}, which was adapted to detect findings in Portuguese for this dataset. CandidPTX~\cite{Feng2021} encompasses data acquired between January 2010 and April 2020 from Dunedin Hospital in New Zealand. This dataset's chest radiographs were manually annotated by RANZCR radiology trainees and radiologists with respect to pneumothoraces, acute rib fractures, and intercostal chest tubes. VinDR~\cite{nguyen2020vinbigdata} was gathered from HMUH and H108 hospitals in Vietnam between 2018 and 2020, with images labeled for six diagnoses by multiple experienced radiologists from these institutions. JF Healthcare~\cite{jfhealthcare} data was collected from approximately 300 township hospitals in China and manually annotated by multiple radiologists to identify foreign objects within the lung field on CXRs. These datasets are comprised of images and associated binary labels that indicate whether common disease entities such as \textit{pneumonia}, or \textit{pneumothorax} are present in the image. However, they lack free-text reports. Thus, to enable pre-training of our image encoder using image and text methods, we create synthetic reports grounded on the labels provided. Detailed templates used for this synthetic rule-based generation can be found in \textbf{Supplementary Table \ref{tab:synthetic}}.

MIMIC-CXR comprises images and their corresponding radiology reports sourced from radiographic studies conducted at the Beth Israel Deaconess Medical Center in Boston, MA, spanning the years 2011 to 2016~\cite{johnson2019mimic}. PadChest~\cite{bustos2020padchest} encompasses CXRs interpreted and reported by 18 radiologists at the Hospital Universitario de San Juan, Alicante (Spain), covering the period from January 2009 to December 2017, alongside their corresponding reports in Spanish. For this data, we harness the capabilities of GPT-4 to translate these reports into English, ensuring linguistic consistency.

\textbf{MIMIC-CXR} free-text reports are utilized for training the text-generation component of \ourmodel. For each report, we extract the \textit{Indication}, \textit{Findings} and \textit{Impression} sections. To do so, we employ rule-based heuristics as supported by the official MIMIC code repository.

Extracting reports in this rule-based manner poses two challenges. First, report structure varies within the dataset, with use of different section headers, merging of findings and impression into the same section, etc., which limits the availability of reports with findings available. Second, reports often contain references to prior examinations, such as "heart size remains unchanged". This poses a challenge for training report generation systems which often hallucinate references to prior examinations that are not available at inference time \cite{ramesh2022improving}. To mitigate these challenges, we leverage GPT-4 to extract the reason for exam, findings, and impression sections in the free-text reports from MIMIC-CXR. Prompt templates used to instruct GPT-4 for the organization are elaborated in \textbf{Supplementary Table \ref{tab:gpt_prompt}}. Compared to the standard MIMIC-CXR rule-based extraction method, GPT-4 demonstrates proficiency in enhancing report quality by addressing issues like grammar errors, broken words, and synonymous section headers, while at the same time eliminating redundant phrases and references to previous exams. \textbf{Supplementary Table \ref{tab:gpt_structured}} showcases examples of sections structured by GPT and those extracted through rule-based methods. The use of GPT to extract sections augments rule-based data by an additional $237,073$ image-text pairs for the training split and $1,952$ for the validation split, as summarized in \textbf{Supplementary Table \ref{tab:mimic_number}}.

\subsection*{Modeling Approach}
\label{sec:modeling_approach}

\textbf{Image Encoder}
Within the LLaVA framework, the image encoder plays a pivotal role in extracting complex image representations, crucial for tasks such as automated medical report generation where standard vision transformer models often do not capture the necessary detail and nuanced representations (\textbf{Supplementary Table \ref{table:llava_evolution}}). To overcome this, we pretrain a domain-specific vision encoder, named \ourvit, and integrate it into LLaVA to bolster its medical image analysis capabilities. Our method includes several key enhancements: firstly, we increase the image input resolution to 518px, substantially higher than the 224px or 336px resolutions typically used in LLaVA-Med, to capture more detailed image features. Secondly, we compile the CXR-697K dataset, an extensive collection of over 697 thousand CXR images from various sources, providing a rich foundation for pretraining. Lastly, we employ the BiomedCLIP recipe for training \ourvit, which involves contrastive vision-language training with PubMedBERT, a text encoder specialized for the medical domain \cite{zhang2024biomedclip}. The initialization of our vision encoder uses a DINOv2 model checkpoint, benefiting from its extensive training on a diverse set of 142 million general-domain images \cite{oquab2024dinov2}.

\textbf{Small Multimodal Model}
\ourmodel leverages the capabilities of a pre-trained image encoder and a pre-trained language model to create a SMM. We choose \ourvit as our image encoder and Vicuna-7B-v1.5~\cite{vicuna2023} as our language model. A multi-layer perceptron (MLP), which is randomly initialized, is introduced to project image features extracted by the image encoder into the word embedding space of the language model. Conditioned on the projected image features (visual tokens) and textual tokens, \ourmodel generates text in an autoregressive manner. We refer the reader to LLaVA~\cite{liu2023llava,liu2023improvedllava} for a more in-depth description of the model architecture.

\textbf{Training Strategy}
Due to the introduction of our domain-specific image encoder \ourvit, \ourmodel is not initialized with the pre-trained LLaVA weights. Instead, we initialize \ourmodel with the pre-trained image encoder \ourvit, the pre-trained language model Vicuna-7B-v1.5, and a randomly initialized MLP.
Similar to LLaVA and LLaVA-Med, the SMM training procedure is done in two stages, feature alignment and end-to-end fine-tuning. Given a set of training examples, where each example consists of a CXR $X_v$ and the corresponding indication section $X_i$ and finding section $X_f$ from the processed report, the training procedure is described as follows: 

In the feature alignment stage, we freeze the image encoder and the language model, and only update the MLP projection layer. Given a CXR $X_v$, we train \ourmodel to generate the corresponding findings section $X_f$. Note that the indication section $X_i$ is not used in this stage. No text prompt is used, and the image is the only input. Our goal is to align the image features with word embeddings of the language model via the learning of the projection layer. In this stage, we train \ourmodel on the training split of MIMIC-CXR for 1 epoch.

In the end-to-end fine-tuning stage, we train both the MLP projection layer and the language model. However, unlike the majority of existing work that fully fine-tunes the language model, we apply the parameter efficient fine-tuning method LoRA~\cite{hu2021lora}, which has recently been shown to achieve comparable performance to full fine-tuning while significantly reducing the training cost~\cite{dettmers2023qlora,lu2023empirical}. 
Given a CXR $X_v$ and the corresponding indication section $X_i$, we train \ourmodel to generate the finding section $X_f$, using the training split of MIMIC-CXR. Similar to the approaches taken by LLaVA and LLaVA-Med, our training process utilizes cross-entropy loss, applied in an auto-regressive manner, to optimize the generation of reports.

\subsection*{Model Evaluation}
\label{sec:evaluation}
Our model evaluation consists of cross-modal retrieval evaluation, where we evaluate the quality of alignment between \ourmodel's CXR their corresponding reports, attention visualization, which illustrates the level of grounding the model's text predictions with regions of the input image, and the automated report evaluation which studied factual correctness and lexical similarity metrics and their alignment with radiologist error quantification. To ensure a thorough evaluation, the model is tested not only on a held-out subset of the MIMIC-CXR dataset, but also on a held-out subset of the CheXpert (n=63) and Open-I~\cite{demner2012design} (n=2,163) datasets. Notably, CheXpert CXRs from the training set were used alongside synthetic label-derived reports to train the image encoder, but the held-out evaluation set, derived from the CheXpert validation split, contains CXRs and radiologist reports that were not available during training. Alternatively, the Open-I dataset was fully held out during model development. The inclusion of CheXpert and Open-I tests the external generalizability and adaptability of the model across different datasets with varying degrees of familiarity and complexity.

\subsubsection*{Image-Text Alignment}
\textbf{Cross-modal retrieval evaluation} This task consists of matching radiology reports to their corresponding radiology images (text-to-image) and the reverse (image-to-text), thus evaluating the model's ability to identify corresponding CXRs and reports by calculating similarity scores between images and text. We compared the performance of \ourmodel, which uses the specialized \ourvit, with more general image encoders used for LLaVA-Med and LLaVA, namely BiomedCLIP and OpenAI CLIP models. We used the official MIMIC-CXR test set for evaluation, quantifying performance using recall at K, a commonly used retrieval evaluation metric that measures the share of relevant items captured within the top K positions.

\textbf{Attention Visualization} 
To qualitatively examine how well \ourmodel's image and text align, we visualize the model's attention mechanisms during its generative process. Specifically, we focus on analyzing \ourmodel's attention to each image token while generating words. This analysis enables us to understand how well each generated word aligns with relevant regions within the image. To achieve this, we conduct an in-depth examination of a fully trained \ourmodel model across all its 32 layers and 32 attention heads. Furthermore, to provide a clearer insight into the model's attention patterns, we calculate either the mean or maximum values (or both) across all layers and heads. For visualization purposes, we reformat the attention matrices into a 37x37 grid to mirror the original spatial dimensions of the image tokens.

\subsubsection*{Quality of Generated Reports}

\textbf{Existing Evaluation Metrics} 
We employ a suite of automatic evaluation metrics to determine the quality of generated reports. We report commonly used lexical similarity metrics (ROUGE-L, BLEU-4) for the sake of comparison with prior methods. However, we focus our model development on factual correctness metrics, employing commonly used metrics such as F1-CheXbert and F1-RadGraph, as well as proposing an automatic GPT scoring-based metric, \ourscore. The F1-CheXbert metric \cite{zhang2020optimizing} corresponds to the F1 score of extracted disease labels of a generated report compared to the reference, as determined by the CheXbert labeler \cite{smit2020chexbert}. In line with prior work, we report F1-CheXbert for all 14 CheXbert classes, in addition to that over 5 classes that represent the most common findings in real-world CXR reports (atelectasis, cardiomegaly, consolidation, edema, and pleural effusion). The F1-RadGraph metric \cite{delbrouck2022improving} broadens the scope of the factual correctness evaluation by comparing the agreement of anatomy and observation entities extracted from the candidate report with that of the reference. Prior to this work, the F1-RadGraph metric was considered as a reference for the evaluation of factual correctness in radiology reports. However, it has limited correlation with manual error scoring as performed by radiologists, which has led to the proposal of composite metrics such as RadCliQ that aim to better reflect human evaluation of factual correctness \cite{yu2023evaluating}.

\textbf{\ourscore}
Given the limitations of existing report evaluation methods and the challenge of accurately evaluating generated reports at scale, we explore the utility of a language model-based scoring system, which has shown success in other domains \cite{liu2023gpteval, wang2023chatgpt, gilardi2023chatgpt}. Specifically, we employ GPT as an evaluator that quantifies the presence of the following six error types, as per \cite{yu2023evaluating}: false prediction of finding, omission of finding, incorrect location/position of finding, incorrect severity of finding, mention of comparison that is not present in the reference, omission of comparison describing a change from a previous study. We instruct GPT to quantify the number of errors of each of the six error types, keeping a separate count for clinically insignificant and significant errors. In each rating prompt, we include a fixed set of five example report evaluation pairs alongside mean error counts for each type to enable the model to leverage in-context examples that quantify errors as requested. We evaluate the validity of the proposed \ourscore using the ReXval dataset \cite{yu2023radiology}, which is comprised of error annotations from 6 board certified radiologists on 200 pairs of candidate and ground-truth reports, where each radiologist provides counts of each of the 6 aforementioned error types, also discriminating between clinically significant and insignificant errors. 

For comparison, we evaluate the performance of three types of GPT models: GPT-3.5 Turbo (GPT-3.5T), GPT-4, and GPT-4-Turbo (GPT-4T). We quantify the alignment between errors quantified by radiologists with that of existing report evaluation methods, in addition to \ourscore, using Kendall's Tau b coefficient (rank correlation coefficient). Further, we directly compare the performance of \ourscore based on GPT-4T with that of each radiologist in a leave-one-rater-out fashion. For each comparison with a rater, the mean of the remaining left-in radiologist raters was calculated. The paired interobserver difference between the held-out radiologist rater and the mean was compared to the paired interobserver difference between \ourscore and the mean. The mean absolute interobserver difference (MAD) for each left-out radiologist was compared with that of \ourscore, with statistical significance determined using paired t-tests. 

Finally, we use the GPT-4T version of \ourscore to quantify the total number of clinically significant and overall errors in each generated report in the evaluation datasets. We quantify these totals in reports from publicly accessible models, enabling us to compare \ourmodel with LLaVA-Med, LLaVA, CheXagent, and GPT-4V. Further, we study the overall proportion of error-free reports in the evaluation datasets, reflecting the potential of each model in directly impacting radiology workflows.

\section*{Data availability}
We will provide scripts to reproduce CXR-697K and MIMIC-CXR from the original datasets, upon publication of this manuscript.

\section*{Code availability}
LLaVA-Rad will be made fully available upon publication, including the model weights and relevant source code for pre-training, fine-tuning, and inference. We will also provide detailed methods and implementation steps to facilitate independent replication. The code to reproduce CheXprompt is publicly available at \url{https://aka.ms/chexprompt}.

\newpage
\bibliography{references}
\bibliographystyle{plain}
\newpage
\newpage

\newpage

\newpage

\newpage

\newpage

\newpage

\newpage
\newpage

\section*{Supplementary Information}

\setcounter{table}{0}
\renewcommand*{\tablename}{Supplementary Table}

\setcounter{figure}{0}
\renewcommand*{\figurename}{Supplementary Figure}

\begin{figure*}[!ht]
\begin{center}
\includegraphics[width=1
\textwidth]{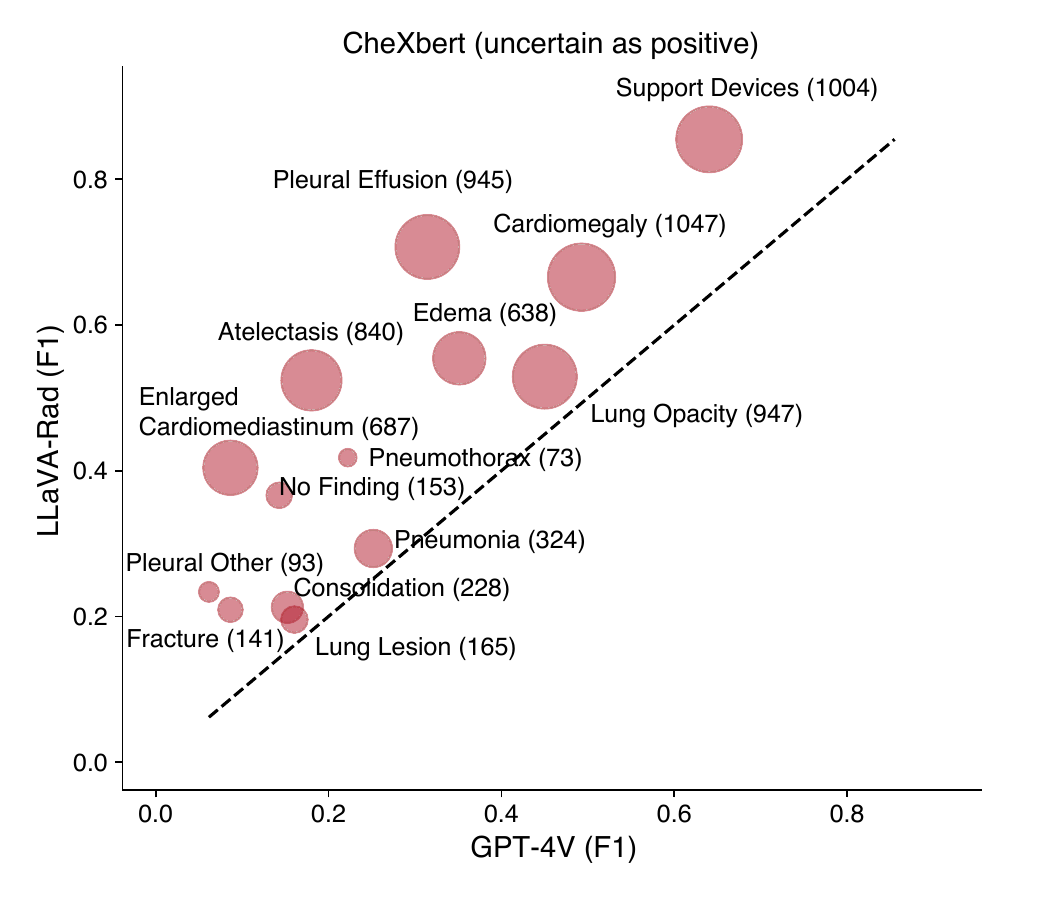}
\end{center}\vspace{-0.01in}
   \caption{Breakdown comparison between \ourmodel{} and GPT-4V on 14 observations extracted by CheXbert from generated reports for the MIMIC-CXR test set. The circle size is proportional to the number of ocurrences of each observation in the reference reports, also provided in parentheses following each observation label.}
\label{fig:breakdown}\vspace{0.05in}
\end{figure*}

\newpage

\begin{figure*}[!ht]
\begin{center}
\includegraphics[width=1
\textwidth]{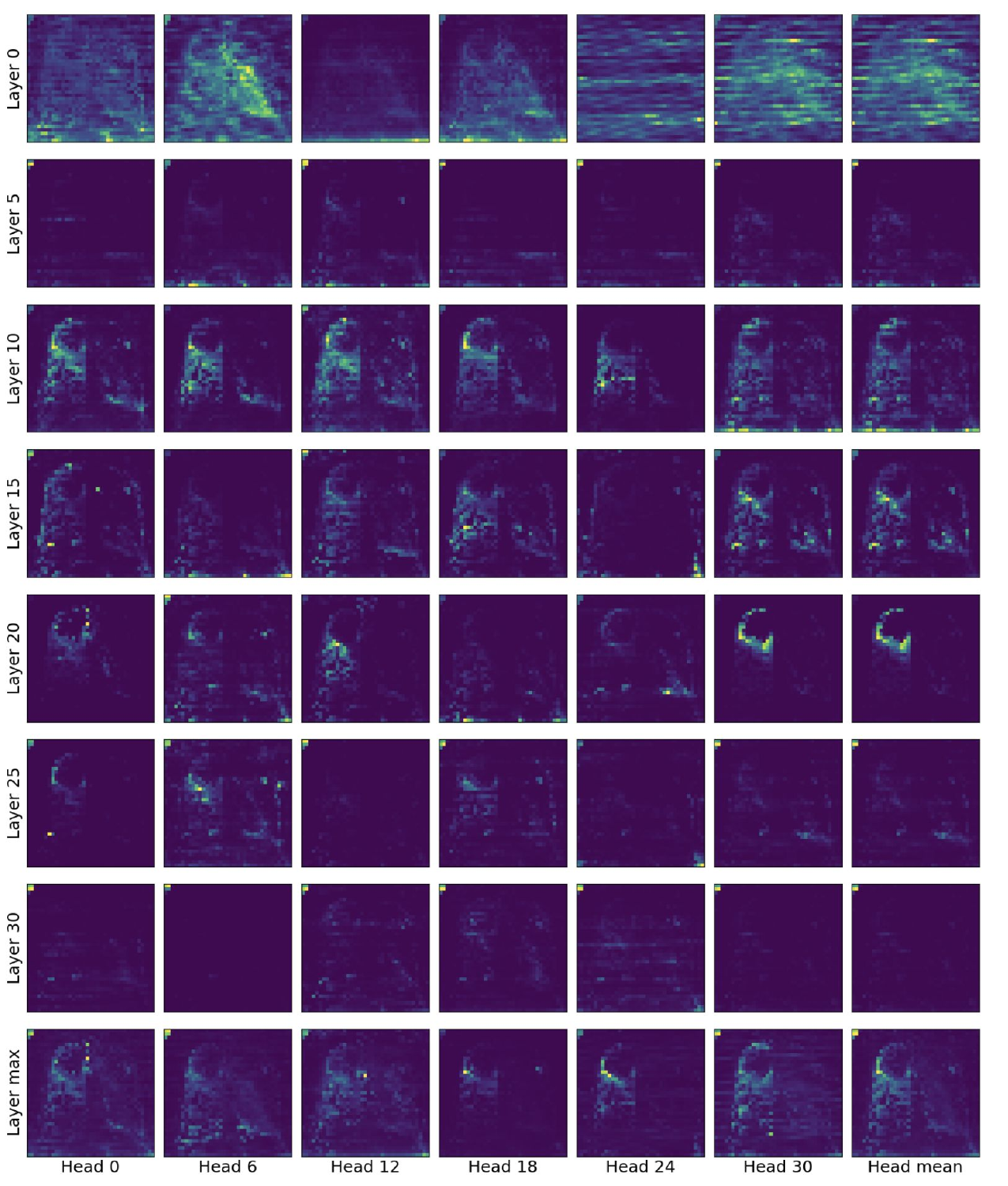}
\end{center}\vspace{-0.01in}
   \caption{Attention weights for the generated word \textbf{Effusion} from different layers and heads of the model.} 
\label{fig:attn-effusion}\vspace{0.05in}
\end{figure*}

\newpage

\begin{figure*}[!ht]
\begin{center}
\includegraphics[width=1
\textwidth]{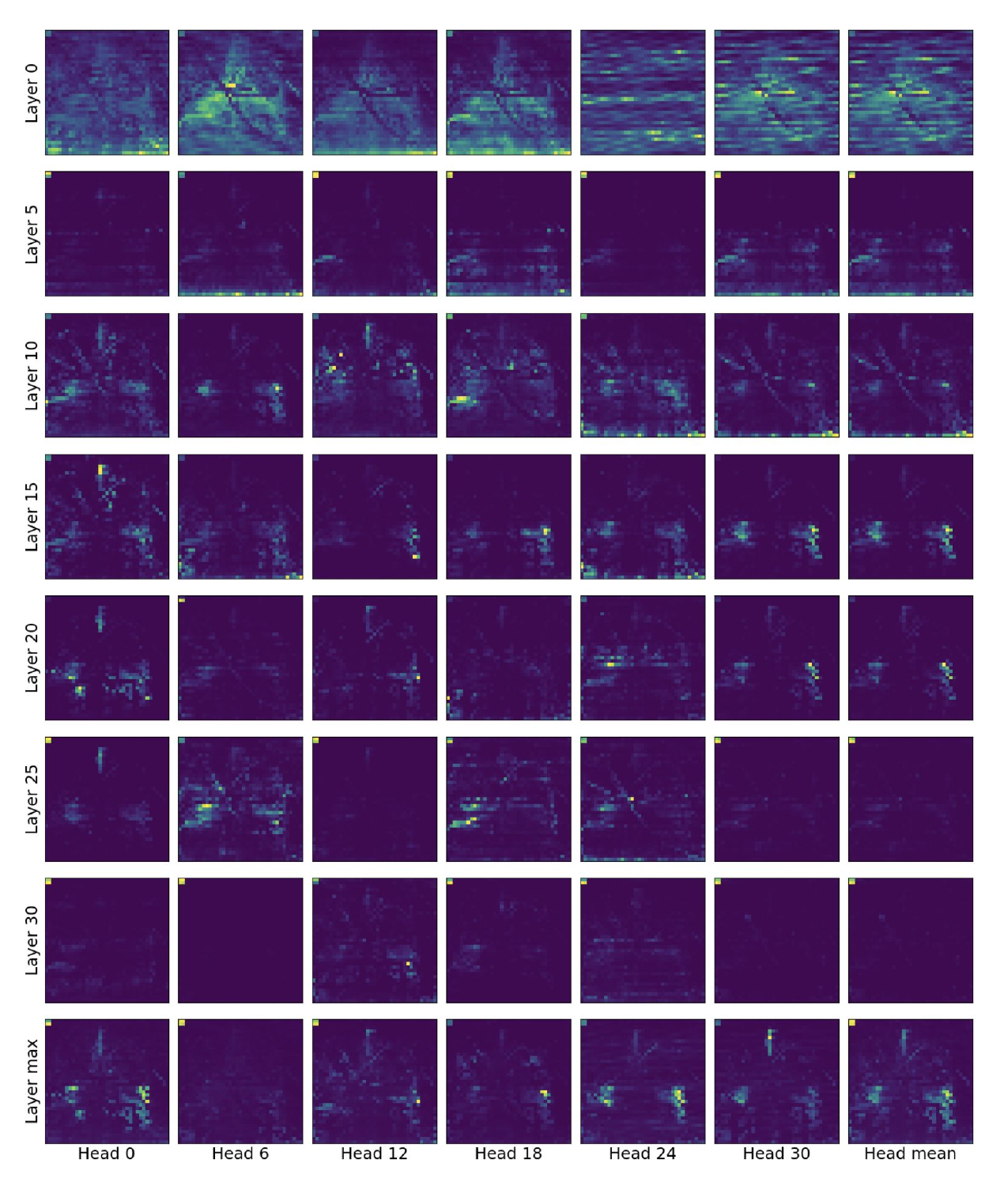}
\end{center}\vspace{-0.01in}
   \caption{Attention weights for the generated word \textbf{Opacification} from different layers and heads of the model.} 
\label{fig:attn-opac}\vspace{0.05in}
\end{figure*}

\newpage

\begin{figure*}[!ht]
\begin{center}
\includegraphics[width=1
\textwidth]{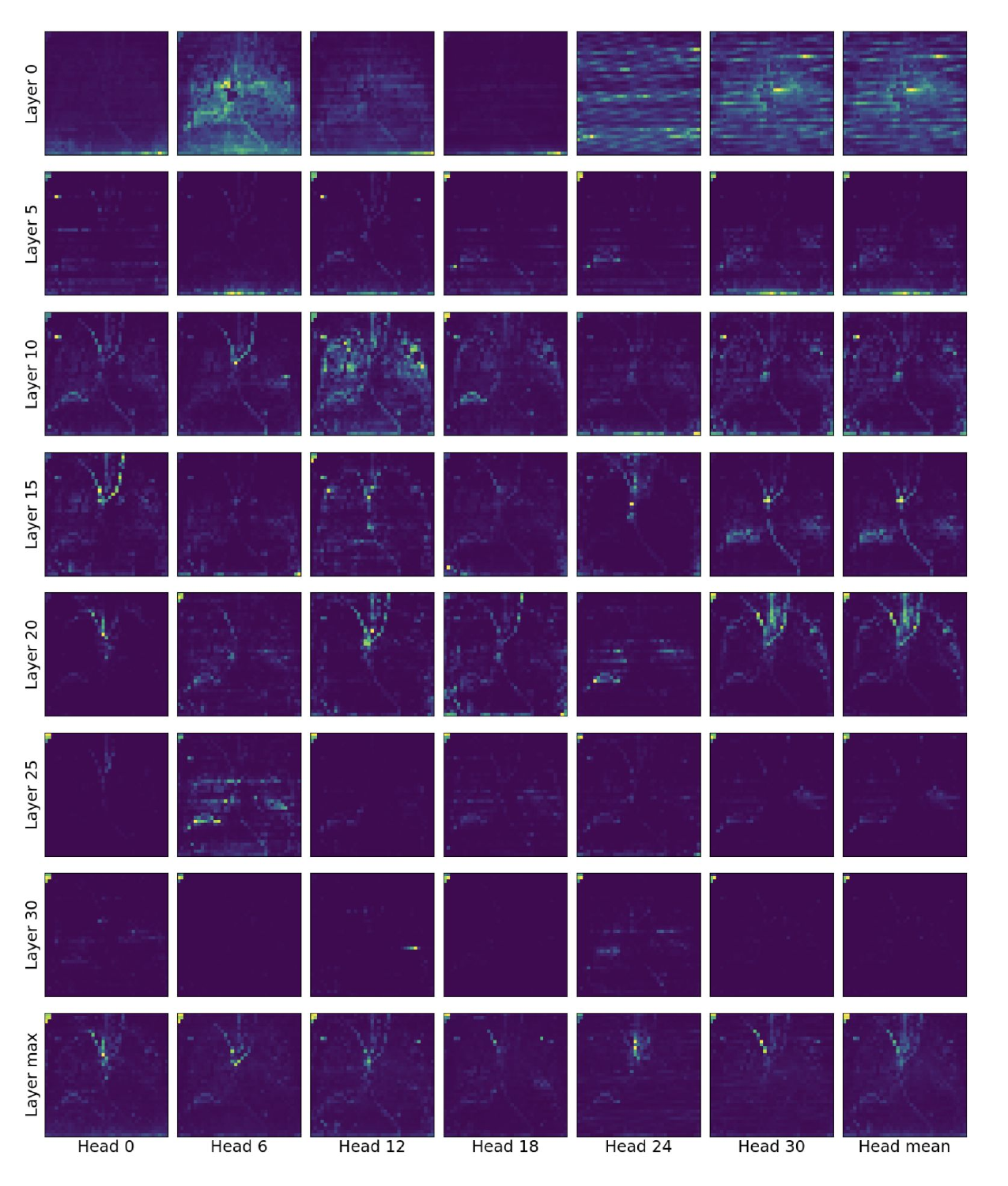}
\end{center}\vspace{-0.01in}
   \caption{Attention weights for the generated word \textbf{devices} from different layers and heads of the model.} 
\label{fig:attn-devices}\vspace{0.05in}
\end{figure*}

\newpage

\begin{figure*}[!ht]
\begin{center}
\includegraphics[width=1
\textwidth]{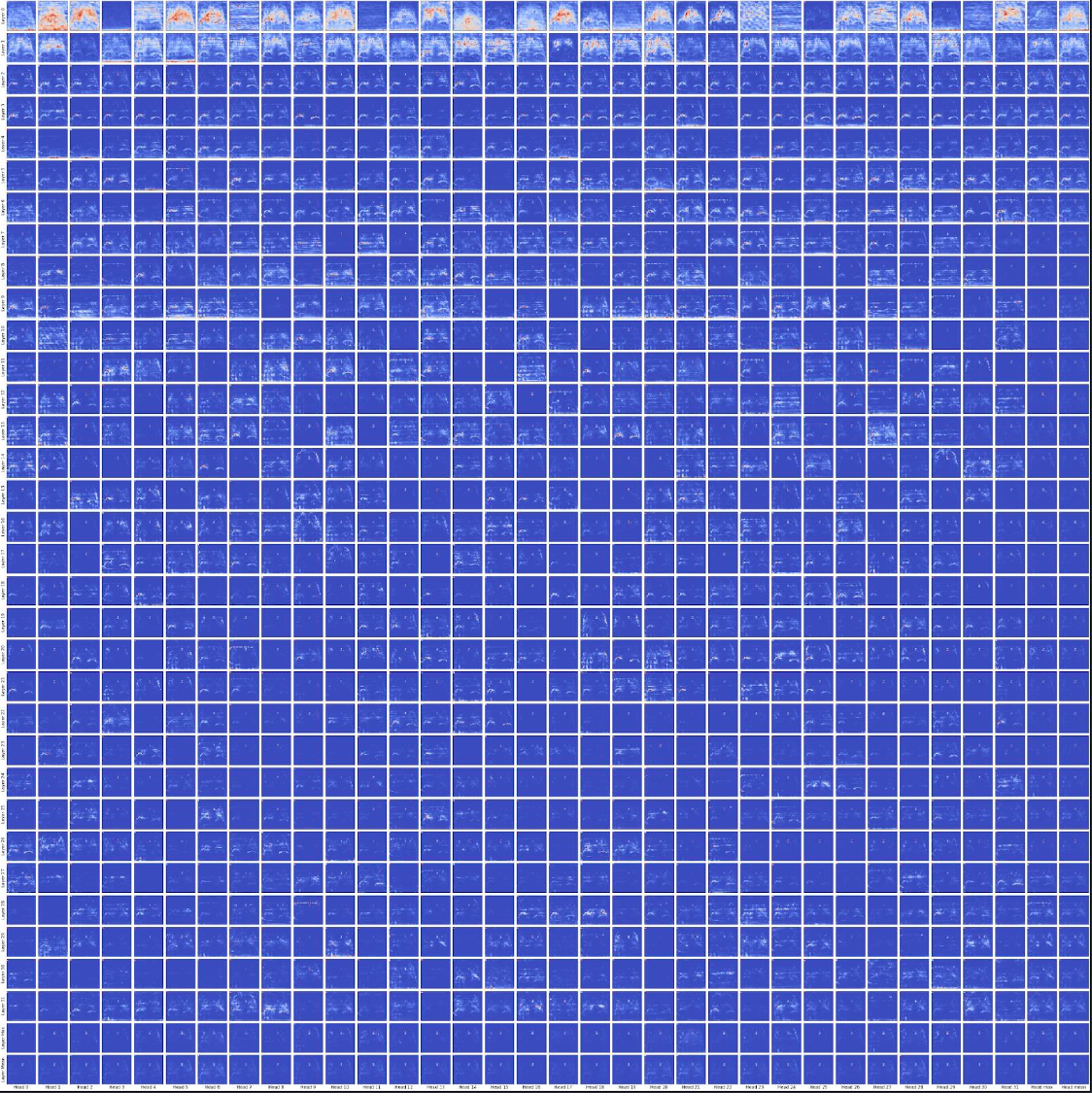}
\end{center}\vspace{-0.01in}
   \caption{All attention weights for the generated word \textbf{aortic} from all layers and heads of the model.} 
\label{fig:attn-aortic}\vspace{0.05in}
\end{figure*}

\newpage
\begin{figure*}[!ht]
\begin{center}
\includegraphics[width=1
\textwidth]{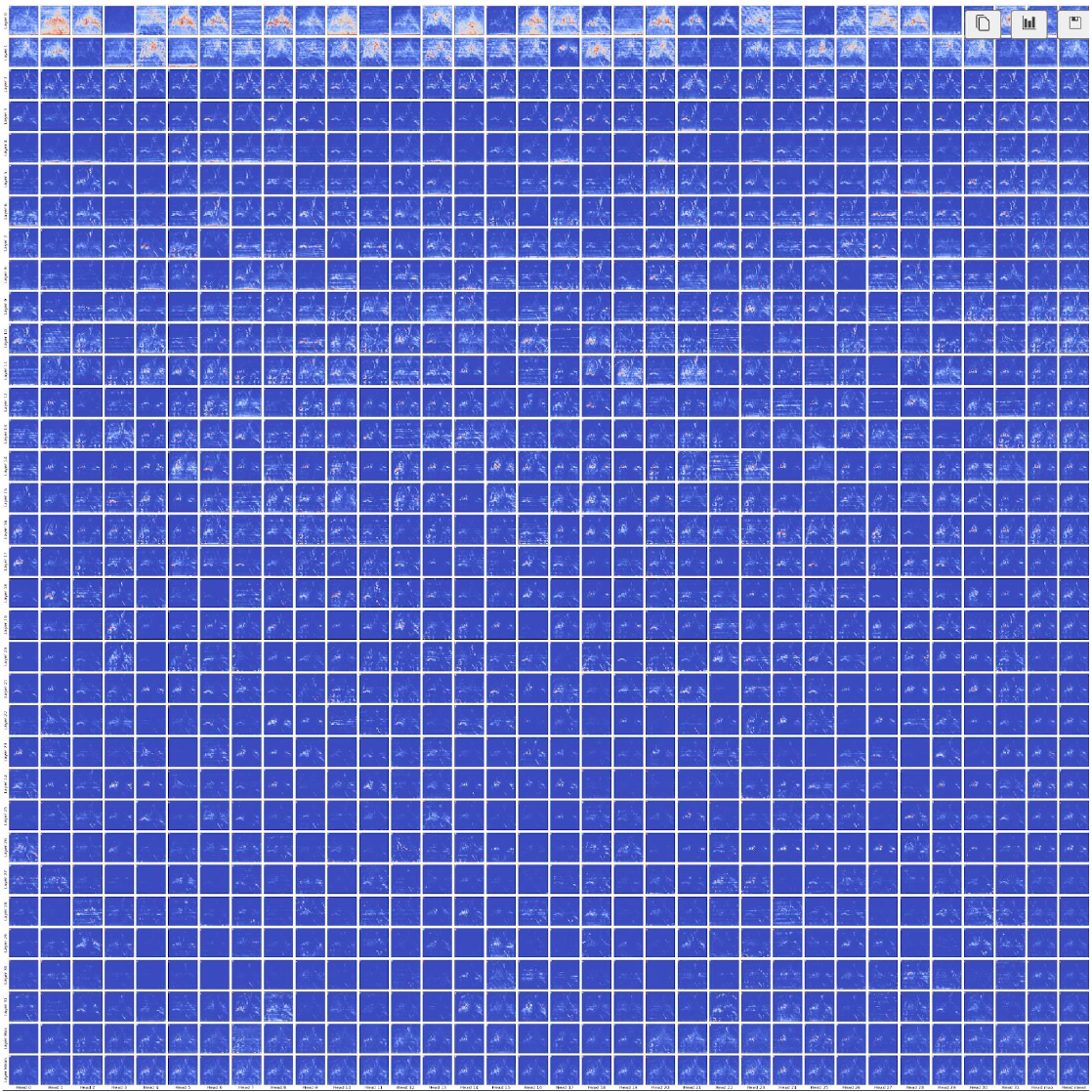}
\end{center}\vspace{-0.01in}
   \caption{All attention weights for the generated word \textbf{opacifications} from all layers and heads of the model.} 
\label{fig:attn-opac-2}\vspace{0.05in}
\end{figure*}

\newpage
\begin{table}[!ht]
\tablestyle{2pt}{1.8}
\begin{tabular}{lcccccccccrrrc} %
\toprule
\multirow{4}{4em}{Model}& \multirow{4}{2em}{Size} & \multicolumn{8}{c}{CheXbert} & \multirow{2}{*}{Rad-} & \multicolumn{2}{c}{\multirow{3}{*}{BLEU}} & \multirow{3}{*}{ROUGE}\\ \cline{3-10}

& & \multicolumn{4}{c}{("uncertain" as \emph{negative})} & \multicolumn{4}{c}{("uncertain" as \emph{positive})} & Graph   & \multicolumn{3}{c}{} \\ \cline{3-10} %
& & \multicolumn{2}{c}{Micro-avg} & \multicolumn{2}{c}{Macro-avg} & \multicolumn{2}{c}{Micro-avg} & \multicolumn{2}{c}{Macro-avg} &  &  & &   \\ %
& & F1-14 & F1-5 & F1-14 & F1-5 & F1-14 & F1-5 & F1-14 & F1-5 & \multicolumn{1}{c}{F1} & \multicolumn{1}{c}{(1)} & \multicolumn{1}{c}{(4)} & \multicolumn{1}{c}{-L}   \\ \shline %

LLaVA-Rad & 7B & \textbf{57.3} & 57.4 & 39.5 & 47.7 & \textbf{57.3} & \textbf{60.2} & \textbf{44.0} & \textbf{53.3} & 29.4 & 38.1 & \textbf{15.4} & \textbf{30.6} \\ %
Med-PaLM M& 84B & 53.6 & \textbf{57.9} & \textbf{39.8} & \textbf{51.6} & - & - & - & - & 26.7 & 32.3 & 11.3 & 27.3 \\ %
GPT-4V & - & 35.5 & 25.8 & 20.4 & 19.6 & 35.6 & 33.3 & 25.3 & 29.6 & 13.2 & 16.4 & 1.9 & 13.2 \\ %
MAIRA-1 & 7B & 55.7 & 56.0 & 38.6 & 47.7 & 55.3 & 58.8 & 42.3 & 51.7 & \textbf{29.6} & 39.2 & 14.2 & 28.9 \\ %
CheXagent & 7B & 39.3 & 41.2 & 24.7 & 34.5  & 39.4 & 42.1 & 27.3 & 35.8 & 20.5 & 16.9 & 4.7 & 21.5 \\ %
LLaVA-Med & 7B & 27.2 & 22.0 & 15.5 & 16.6 & 27.3 & 24.4 & 18.7 & 20.5 & 6.5 & 22.2 & 1.0 & 13.3 \\ %
LLaVA & 7B & 22.9  & 23.4 & 15.4 & 17.5 & 23.7 & 26.9 & 17.0 & 20.3 & 4.5 & 21.0 & 1.3 & 13.8 \\ %
Flamingo-CXR & <1B & - & - & - & - & 51.9 & 56.5 & - & - & 20.5 & - & 10.1 & 29.7 \\ %
CvT2Dist. & <1B & 44.2 & - & 30.7 & - & - & - & - & - & - & \textbf{39.3} & 12.7 & 28.6 \\ %
$\mathcal{M}^2$ trans & <1B & - & - & - & - & - & 56.7 & - & - & - & - & 11.4 & - \\ %
RGRG & <1B & - & - & - & - & - & 54.7 & - & - & - & 37.3 & 12.6 & 26.4 \\ %
R2Gen & <1B & - & - & - & - & 22.8 & 34.6 & - & - & - & 35.3 & 8.6 & 27.7 \\ %
TieNet & <1B & - & - & - & - & - & 27.1 & - & - & - & - & 8.1 & - \\ %
\bottomrule
\end{tabular}
\vspace{3mm}
\caption{Evaluation results on the MIMIC-CXR test set. B denotes billions in the column of Size.}
\label{tab:main-results}
\end{table}

\begin{table}[!ht]
\tablestyle{2pt}{1.8}
\begin{tabular}{lcccccccccccc}
\toprule
\multirow{4}{4em}{Model Version} & \multicolumn{8}{c}{CheXbert} & \multirow{2}{*}{Rad-} & \multicolumn{2}{c}{\multirow{3}{*}{BLEU}} & \multirow{3}{*}{ROUGE}\\ \cline{2-9}
& \multicolumn{4}{c}{("uncertain" as \emph{negative})} & \multicolumn{4}{c}{("uncertain" as \emph{positive})} & Graph   & \multicolumn{3}{c}{} \\ \cline{2-9} %
& \multicolumn{2}{c}{Micro-avg} & \multicolumn{2}{c}{Macro-avg} & \multicolumn{2}{c}{Micro-avg} & \multicolumn{2}{c}{Macro-avg} &  &  &   \\ 
& F1-14 & F1-5 & F1-14 & F1-5 & F1-14 & F1-5 & F1-14 & F1-5 & F1 & (1) & (4) & -L \\ \shline 
\ourmodel & \textbf{57.3} & \textbf{57.4} & \textbf{39.5} & 47.7 & \textbf{57.3} & \textbf{60.2} & \textbf{44.0} & 53.3 & 29.4 & 38.1 & \textbf{15.4} & \textbf{30.6} \\ \hline
Variant \#1 & 55.5 & 55.2 & 38.4 & 46.9 & 55.9 & 58.4 & 42.5 & 51.1 & 29.6 & \textbf{38.5} & \textbf{15.4} & \textbf{30.6} \\
Variant \#2 & 55.6 & 56.4 & 38.5 & \textbf{48.6} & 55.7 & 59.5 & 41.6 & 52.9 & 28.9 & 37.5 & 15.2 & 30.3 \\ 
Variant \#3 & 55.7 & 56.2 & 39.1 & 47.4 & 55.8 & 59.0 & 43.1 & 51.4 & \textbf{29.7} & 37.9 & 15.6 & 31.0 \\ 
Variant \#4 & 56.2 & 56.9 & 38.6 & 48.3 & 56.1 & \textbf{60.2} & 43.0 & \textbf{54.0} & 27.6 & 29.9 & 10.0 & 26.3 \\ 
Variant \#5 & 50.6 & 53.3 & 30.7 & 42.7 & 49.2 & 54.1 & 33.6 & 45.0 & 25.0 & 16.9 & 5.0 & 16.4 \\
Variant \#6 & 51.1 & 52.9 & 31.0 & 43.6 & 51.4 & 56.3 & 35.1 & 48.6 & 26.9 & 35.0 & 13.9 & 29.4 \\
Variant \#7 & 49.8 & 52.1 & 30.5 & 42.4 & 50.3 & 55.4 & 34.5 & 48.0 & 26.5 & 34.4 & 13.7 & 29.1 \\
Variant \#8 & 49.6 & 52.2 & 31.3 & 42.8 & 50.1 & 55.6 & 35.0 & 47.8 & 26.9 & 34.3 & 13.7 & 29.3 \\ \hline 
Analysis \#1 & 96.7 & 98.3 & 95.1 & 97.8 & 96.9 & 98.4 & 95.5 & 97.8 & 93.2 & 82.2 & 78.8 & 90.9 \\ \shline
\multicolumn{13}{l}{\textbf{Stage 1} (\emph{image encoder pre-training}) $\rightarrow$ \textbf{Stage 2} (\emph{alignment}) $\rightarrow$ \textbf{Stage 3} (\emph{fine-tuning})}\\
\multicolumn{13}{l}{\quad Variant \#1: Stage 1 pre-trains the image encoder on MIMIC-CXR only. Stage 2 and 3 are the same.}\\ 
\multicolumn{13}{l}{\quad Variant \#2: No stage-1 pretraining. Stage 2 and 3 are the same.}\\  %
\multicolumn{13}{l}{\quad Variant \#3: Stage 1 is the same. Stage 2 and 3 only use rule-processed MIMIC-CXR training data.} \\ 
\multicolumn{13}{l}{\quad Variant \#4: Stage 1 is the same. Stage 2 and 3 only use GPT-4 processed MIMIC-CXR training data.} \\ 
\multicolumn{13}{l}{\quad Variant \#5: Stage 1 and 2 are the same. No stage 3.}\\
\multicolumn{13}{l}{\quad Variant \#6: No stage 1. Stage 2 initializes with LLaVA-v1.5 pre-trained weights.}\\
\multicolumn{13}{l}{\quad Variant \#7: No stage 1. Stage 2 initializes the image encoder with OpenAI CLIP pre-trained weights.}\\
\multicolumn{13}{l}{\quad Variant \#8: No stage 1. Stage 2 initializes the image encoder with BiomedCLIP pre-trained weights.}\\
\multicolumn{13}{l}{\quad Analysis \#1: Evaluate GPT-4 processed test data against rule-processed test data.} \\ \shline
\end{tabular}
\vspace{3mm}
\caption{Ablation study.}
\label{tab:ablation}
\end{table}

\newpage
\begin{table}[!ht]
\centering
\begin{tabular}{lrrr}
\toprule
 MIMIC-CXR & Training & Validation & Test \\
 \midrule
 Patients & $63,169$ & $487$ & $289$\\
 Studies & $213,365$ & $1,733$ & $3,041$\\
DICOMs (AP/PA) & $237,972$ & $1,959$ & $3,403$ \\
\midrule
Rule-based reports & $162,969$ & $1,286$ & $2,461$ \\
GPT-structured reports & $237,073$ & $1,952$ & - \\
\midrule
Total data & $400,042$ & $3,238$ & $2,461$ \\
\bottomrule
\end{tabular}
\vspace{3mm}
\caption{Number of patients, studies and DICOMs in AP/PA views across official MIMIC-CXR splits. Numbers of image-text pairs generated by official rule-based method and augmented by GPT-4 in training and validation splits, excluding texts with no \textit{Findings} extracted.}
\label{tab:mimic_number}
\end{table}

\newpage 

\begin{table}[ht]
\centering
\tablestyle{7pt}{1.3}
\begin{tabular}{lrrrrrrr}
\toprule[1.5pt]
Dataset & Patients & Images  & \multicolumn{1}{c}{Label} & \multicolumn{3}{c}{Text Length} & \multicolumn{1}{c}{Pre-training} \\ \cline{5-7}
& & & \multicolumn{1}{c}{Types} & Min & Max & Avg. & \multicolumn{1}{c}{Image-Text Pairs} \\
\midrule[1pt]
& & & & \multicolumn{3}{c}{\textit{Synthetic Findings}} & \\
CheXpert~\cite{irvin2019chexpert} & $64,540$ & $224,316$ & $14$ & $3$ & $89$ & $27.8$  & $190,999$\\
BraX~\cite{reis2022brax} & $19,351$ & $40,967$ & $14$ & $3$ & $76$ & $32.1$ & $19,309$ \\
CandidPTX~\cite{Feng2021} & $13,744$ & $19,237$ & $3$ & $12$ & $33$ & $22.8$ & $19,237$ \\
VinDR~\cite{nguyen2020vinbigdata} & $18,000$ & $18,000$ & $6$ & $79$ & $125$ & $103.2$ & $15,000$  \\
JF Healthcare \cite{jfhealthcare} & $10,000$ & $10,000$  & $1$ & $4$ & $12$ & $7.6$ & $10,000$ \\
\midrule
& & & & \multicolumn{3}{c}{\textit{Real-World Findings}} & \\
MIMIC-CXR~\cite{johnson2019mimic} & $65,379$ & $377,095$ &  -- & $3$ & $183$ & $85.1$ & $353,350$ \\
PadChest~\cite{bustos2020padchest} & $67,625$ & $168,861$ & -- & $1$ & $133$ & $12.5$ & $89,540$ \\ \cline{8-8}
& & & & & & & Total: $697,435$\\
\bottomrule[2.0pt]
\end{tabular}

\vspace{3mm}

\caption{An extensive collection of 697 thousand Chest X-ray datasets used for pre-training a domain-specific image encoder \ourvit{}. MIMIC-CXR contains images with both frontal and lateral views, while others only include frontal views. \textit{Synthetic Findings} are generated via template sentences derived from supervised clinical labels; \textit{Real-World Findings} are extracted from patient reports utilizing GPT.}
\label{table:dataset}
\end{table}

\begin{table}[!ht]
    \centering
    \begin{tabular}{c}
    \toprule
    \begin{minipage}{\textwidth}
    \small
      \begin{verbatim}
Positive <condition>: 
1. The radiograph reveals evidence of <condition>. 
2. The radiograph demonstrates areas consistent with <condition>. 
3. There are findings suggestive of <condition>. 
4. There is presence of <condition>. 
5. There is a positive finding of <condition>. 
6. The radiographic examination of the chest reveals the presence of <condition>. 
7. There is evidence of <condition>. 
8. <Condition> is present. 

Negative <condition>: 
1. No evidence of <condition> is observed. 
2. <Condition> is not identified. 
3. The radiograph does not show any signs of <condition>. 
4. There is no indications of <condition> in the radiograph. 
5. No <condition> is identified in the examined region. 
6. No signs of <condition> is observed. 
7. The image does not conclusively indicate <condition>. 

Uncertain <condition>: 
1. There is uncertainty regarding <condition>. 
2. The presence of <condition> is uncertain based on the current examination. 
3. The image shows uncertainty in <condition>. 

No findings: 
1. The radiographic examination of the chest reveals no significant abnormalities 
   or pathologies. 
2. The radiographic examination of the patient does not reveal any significant 
   abnormal findings. 

No support devices: 
1. There is no evidence of any support devices in the chest area. 
2. There are no support devices seen in the current study. 
3. There are no support devices in place. 

Pleural others: 
1. There are some pleural abnormalities that do not fit into the common categories 
   of pleural diseases. 
2. Other pleural abnormalities are also observed. 

No pleural others: 
1. No other pleural abnormalities were detected in the radiograph. 
2. There are no findings related to other pleural abnormalities. 

Uncertain pleural others: 
1. There are ambiguous findings related to the pleura. 
2. There is no visible pleural abnormality, although the image does not completely 
   exclude all potential pleural conditions. 
\end{verbatim}
\end{minipage} \\
\bottomrule
    \end{tabular}
    \vspace{2mm}
    \caption{Template sentences for deriving synthetic reports from supervised clinical conditions.}
    \label{tab:synthetic}
\end{table}

\newpage
\begin{table}[!ht]
    \centering
    \begin{tabular}{c}
    \toprule
    \begin{minipage}{\textwidth}
    \small
      \begin{verbatim}
You are an expert medical assistant AI capable of modifying clinical documents to 
user specifications. You make minimal changes to the original document to satisfy 
user requests. You never add information that is not already directly stated in 
the original document.   
      
Extract four sections from the input radiology report: `Examination', `Indication', 
`Findings' and `Impression'. Leave an extracted section as null if it does not 
exist in the original report. The output should be in JSON format. An Indication 
section can refer to the History, Indication or Reason for Study sections in the 
original report. Remove any information not directly observable from the current 
imaging study. For instance, remove any patient demographic data, past medical 
history, or comparison to prior images or studies. The generated `Findings' and 
`Impression' sections should not reference any changes based on prior images, 
studies, or external knowledge about the patient. Rewrite such comparisons as a 
status observation based only on the current image or study. Remember to remove 
any numbering or bullets.

Examples of inputs and expected outputs:

INPUT:
EXAMINATION: XR CHEST AP PORTABLE
INDICATION: Small right apical pneumothorax after lung biopsy.
FINDINGS: Single portable view of the chest was obtained. Copared with 10:42 AM. 
The small right apical pneumothorax has decreased slightly in size, the improvement 
best appreciated laterally where it now measures 10 mm compared to 14 mm before. At 
the lung apex it now measures 1.6 and compared to 2.1 cm previously. A subtle right 
apical pulmonary contusion is grossly stable. Minor chest wall emphysema along the 
right exilla has not changed significant delay. There is no metastatic shift. No 
pleural effusion is evident.

OUTPUT:
{"EXAMINATION": "XR CHEST AP PORTABLE.",
"INDICATION": "Small right apical pneumothorax after lung biopsy.",
"FINDINGS": "Single portable view of the chest was obtained. The small right apical
pneumothorax measures 10mm. At the lung apex it measures 1.6cm. A subtle right apical 
pulmonary contusion is grossly stable. Minor chest wall emphysema is noted along the 
right exilla. There is no metastatic shift. No pleural effusion is evident.",
"IMPRESSION": null}  
\end{verbatim}
\end{minipage} \\
\bottomrule
    \end{tabular}
    \vspace{2mm}
    \caption{GPT Prompt for report structuring.}
    \label{tab:gpt_prompt}
\end{table}

\newpage
\begin{table}[ht]
\tablestyle{6pt}{1.8}
    \centering
    \begin{tabular}{l p{11cm}}
    \toprule
        Original &   INDICATION: \_\_\_ year old woman with likely ileus after cystectomy  //  NGT placement confirmation      NGT placement confirmation

 IMPRESSION: No previous images.  Nasogastric tube extends to the mid body of the stomach,
 be for coiling on itself so that the tip lies close to the esophagogastric
 junction. For more optimal positioning, the to would have to be pulled back
 almost 10 cm and then hopefully redirected toward the lower stomach.
 Cardiac silhouette is within normal limits and there is no vascular
 congestion, pleural effusion, or acute focal pneumonia.\\
    \midrule
    Rule-based &  \textbf{INDICATION}:  \_\_\_ year old woman with likely ileus after cystectomy  // \colorbox{Apricot}{NGT placement confirmation      NGT placement confirmation}
 
\textbf{FINDINGS}:  \colorbox{GreenYellow}{nan}
 
 \textbf{IMPRESSION}: \colorbox{Salmon}{No previous images.}  Nasogastric tube extends to the mid body of the stomach,
 \colorbox{Aquamarine}{be for} coiling on itself so that the tip lies close to the esophagogastric
 junction.  For more optimal positioning, the \colorbox{Aquamarine}{to} would have to be pulled back
 almost 10 cm and then hopefully redirected toward the lower stomach.
 Cardiac silhouette is within normal limits and there is no vascular
 congestion, pleural effusion, or acute focal pneumonia.\\
    \midrule
        GPT-structured &  \textbf{INDICATION}: \_\_\_ year old woman with likely ileus after cystectomy  // \colorbox{Apricot}{NGT placement confirmation}
 
 \textbf{FINDINGS}: \colorbox{GreenYellow}{Nasogastric tube extends to the mid body of the stomach,} \colorbox{Aquamarine}{before} \colorbox{ GreenYellow}{coiling on itself so that the tip lies close to the esophagogastric junction.}  \colorbox{GreenYellow}{Cardiac silhouette is within normal limits and there is no vascular congestion, } \colorbox{GreenYellow}{pleural effusion, or acute focal pneumonia.}
 
 \textbf{IMPRESSION}: For more optimal positioning, the \colorbox{Aquamarine}{tube} would have to be pulled back almost 10 cm and then hopefully redirected toward the lower stomach.\\
    \bottomrule
    \end{tabular}
    \vspace{3mm}
    \caption{Comparison between GPT-structured sections and rule-based extractions. GPT affixes broken words (in \colorbox{Aquamarine}{aquamarine}), removes repeated phrases (in \colorbox{Apricot}{apricot}) and references to prior images (in \colorbox{Salmon}{salmon}), and arranges the text into appropriate \textit{Findings} and \textit{Impression} sections (in \colorbox{GreenYellow}{green yellow}) without being constrained by the original section title.}
    \label{tab:gpt_structured}
\end{table}

\newpage

\begin{table}[!ht]
\centering
\begin{tabularx}{\textwidth}{llll}
\toprule[1pt]
\textbf{Model} & \textbf{Key Training Data} & \textbf{Image Encoder} & \textbf{Key Evaluations} \\ 
\midrule[0.7pt]
LLaVA & Image Captions + GPT4 & CLIP & VQA, LLaVA-Bench \\ 
LLaVA-Med & PMC Captions + GPT4 & BiomedCLIP & VQA, LLaVA-Bench \\ 
LLaVA-Rad & Radiology Data + GPT4 & \ourvit & \ourscore/CheXpert \\ 
 \bottomrule[1pt]
\end{tabularx}
\vspace{3mm}
\caption{Evolution of the LLaVA model families for biomedical domains}
\label{table:llava_evolution}
\end{table}

\end{document}